\documentclass{article} 
\usepackage[final]{colm2026_conference}
\usepackage{subcaption}
 
\usepackage{booktabs}
\usepackage{ulem}
\usepackage[accsupp]{axessibility}  
\usepackage{microtype}
\usepackage{url}

\usepackage{lineno}

\usepackage{hyperref}

\usepackage{orcidlink}

\usepackage[most]{tcolorbox}
\usepackage[table]{xcolor}

\usepackage{gensymb}

\newcommand{\gemini}{Gemini-2.5-Pro}
\newcommand{\gpt}{GPT-5.2}

\newcommand{\clipL}{CLIP ViT-L/14-336}
\newcommand{\dino}{DINOv2 ViT-L/14}

\newcommand{\siglip}{SigLIP-SO400M-384}
\newcommand{\qwenTwo}{Qwen2.5-VL}
\newcommand{\qwenThree}{Qwen3-VL}
\newcommand{\qwenSeven}{Qwen2.5-VL-7B}
\newcommand{\qwenThirtyTwo}{Qwen2.5-VL-32B}
\newcommand{\qwenSeventyTwo}{Qwen2.5-VL-72B}
\newcommand{\qwenThreeFour}{Qwen3-VL-4B}
\newcommand{\qwenThreeEight}{Qwen3-VL-8B}
\newcommand{\qwenThreeThirty}{Qwen3-VL-30B}

\definecolor{scriptteal}{RGB}{23,190,207}
\definecolor{scriptorange}{RGB}{230,145,75}
\definecolor{scriptpurple}{RGB}{148,103,189}
\definecolor{scriptviolet}{RGB}{120,110,170}
\definecolor{scriptindigo}{RGB}{75,0,130}
\definecolor{scriptgold}{RGB}{188,158,33}
\definecolor{scriptone}{RGB}{166,97,26}   
\definecolor{scripttwo}{RGB}{223,194,125} 
\definecolor{scriptthree}{RGB}{128,115,172} 

\newcommand{\up}{\,\textcolor{teal}{\(\uparrow\)}}
\newcommand{\down}{\,\textcolor{red}{\(\downarrow\)}}
\newcommand{\same}{\,\phantom{\textcolor{teal}{\(\uparrow\)}}}
\newcommand{\tprhead}{TPR\same}

\newcommand{\best}[1]{\textcolor{teal}{#1}}

\usepackage{xcolor}
\definecolor{shadecolor}{RGB}{255,255,200}
\usepackage{tcolorbox}
\usepackage{enumitem}
\definecolor{scriptblue}{RGB}{31,119,180}   
\definecolor{scriptgreen}{RGB}{44,160,44}   
\definecolor{scriptred}{RGB}{214,39,40}     
\usepackage{booktabs}
\usepackage[font=scriptsize]{caption}
\usepackage{array}
\usepackage{arydshln}
\usepackage{adjustbox}
\usepackage{multirow}

\usepackage[most]{tcolorbox}
\tcbuselibrary{skins,breakable,raster}
\usepackage{graphicx}

\usepackage{arydshln}

\title{Semantic Richness or Geometric Reasoning? The Fragility of VLM's Visual Invariance}

\author{
Jason Qiu$^{1*}$\quad\quad
Zachary Meurer$^{1*}$\quad\quad
Xavier Thomas$^{1*\dagger}$\quad\quad
Deepti Ghadiyaram$^{1}$ \\
$^1$Boston University \quad \\
{\tt\small \{jasonq, zmeurer, xthomas, dghadiya\}@bu.edu} \\
{\small $^*$Equal contribution.\quad $^\dagger$Corresponding author.}
}

\begin{document}

\ifcolmsubmission
\linenumbers
\fi

\begin{figure}[!t]
\maketitle
  \centering
  \includegraphics[width=0.8\linewidth, keepaspectratio]{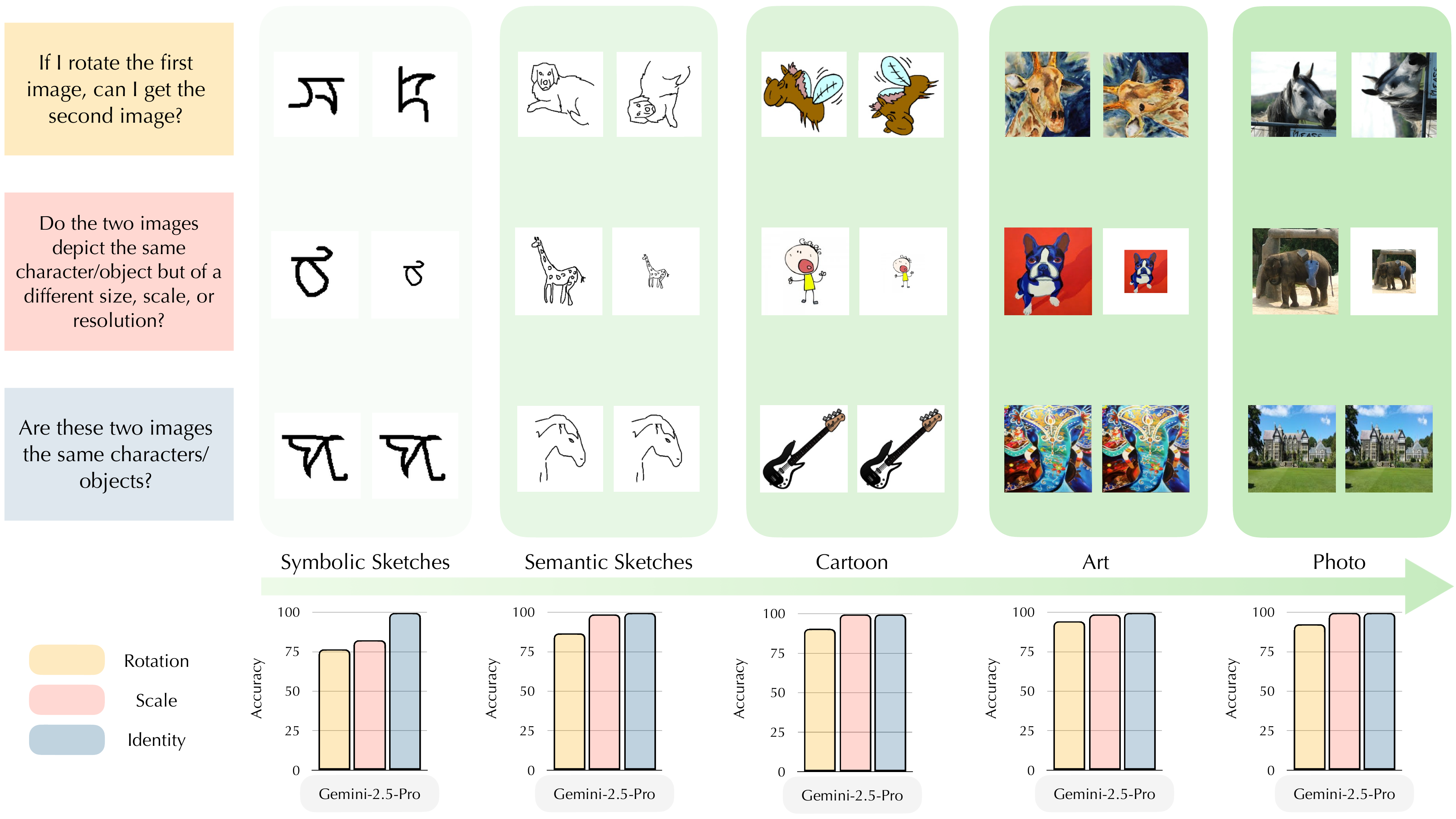}
\vspace{-0.10in}
\caption{
\textbf{Failure of visual transformation reasoning across visual domains.} Given a pair of images, models are asked to determine whether they depict the same object under transformations of rotation, scale, or identity. While performance remains near-perfect on natural images (Art, Photo), accuracy drops sharply on abstract and symbolic images (Symbolic and Semantic Sketches), particularly for rotation. Results shown are for Gemini-2.5-Pro~\citep{gemini}, with similar trends across evaluated MLLMs.}
\vspace{-0.15in}
  \label{fig:main}
\end{figure}

\begin{abstract}
This work investigates the fundamental fragility of state-of-the-art Vision-Language Models (VLMs) under basic geometric transformations. While modern VLMs excel at semantic tasks such as recognizing objects in canonical orientations and describing complex scenes, they exhibit systematic failures at a more fundamental level: lack of robust spatial invariance and equivariance required to reliably determine object identity under simple rotations, scaling, and identity transformations. We demonstrate this limitation through a systematic evaluation across diverse visual domains, including symbolic sketches, natural photographs, and abstract art. Performance drops sharply as semantic content becomes sparse, and this behavior is observed across architectures, model capacities, and prompting strategies. Overall, our results reveal a systematic gap between semantic understanding and spatial reasoning in current VLMs, highlighting the need for stronger geometric grounding in future multimodal systems. Code is available at: \href{https://xthomasbu.github.io/visual_invariance/}{xthomasbu.github.io/visual\_invariance}
\end{abstract}

\section{Introduction}
\label{sec:intro}
Imagine being presented with a sentence in an unfamiliar script, such as \textit{Glagolitic}. To identify repeating characters, we tend to rely solely on rigorous geometric analysis -- matching curves, angles, and topology of characters. Now, consider performing the same task on your native script. The task becomes almost trivial due to semantic familiarity with the characters bypassing the need for shape reasoning. This human ability to fluidly switch between geometric reasoning and semantic recognition as needed raises a critical question: do present day vision-language models (VLMs) possess similar robustness? 

To study this, we evaluate models across a spectrum of semantic granularity, ranging from sparse symbolic sketches and handwritten scripts to texture-rich photographs (Figure~\ref{fig:main}). Within these domains, we test three fundamental transformations: rotation, scaling, and identity matching. A VLM truly possessing geometric reasoning should identify if two images depict the same content regardless of the transformation applied.  Crucially, this capability should remain consistent across both familiar (e.g., Latin) and unfamiliar (e.g., Glagolitic) scripts, semantic sketches, and real photos, as the underlying reasoning remains identical. If, however, a model’s apparent robustness is merely a byproduct of data familiarity or context, the performance should collapse on semantically sparser content. 

Our in-depth analysis confirms this suspicion: the performance of even top-tier closed- and open-sourced VLMs collapses on semantically sparse content under basic geometric transformations. We demonstrate that the apparent invariance observed in real-world images is not just because of geometric reasoning, but rather a byproduct of dataset familiarity and prompt sensitivity.
Specifically, models achieve higher accuracy when asked if two images contain the \textit{same object} than when asked if one is a \textit{rotated variant} of the other. This inherent reliance on object labels -- without a corresponding grasp of the object’s underlying geometry -- reveals a fundamental flaw in current VLMs. Given their pervasive deployment in safety-critical fields like robotics~\citep{li2024manipllm}, this inability to decouple semantic identity from geometric orientation poses a significant risk to reliable spatial interaction.

While prior research has documented VLM failures on high-level reasoning tasks such as visual analogical reasoning~\citep{yiu2024kiva}, object instance orientation~\citep{telling_left_right}, depth estimation~\citep{hemmat2024hidden}, or spatial correspondence~\citep{visually_prompted}, we go beyond identifying failure modes.  By using the gradient of semantic information (hand written digits, sketches, and cartoons) as a ``stress test'' for pure shape and geometric perception, we determine if VLMs rely on universal geometric principles or semantic familiarity. Our results reveal several critical gaps of VLMs, which we summarize below:
\begin{itemize}
    \item All models are highly sensitive to \textbf{data familiarity} in the case of symbolic scripts. They exhibit high performance on familiar scripts such as \textit{Latin} and substantially lower performance on less familiar scripts such as \textit{Grantha}.
    
    \item All models \textbf{perform best on real photos}, remain relatively strong on cartoons, but \textbf{degrade sharply on sketches and symbolic scripts }(Fig.~\ref{fig:main}), where semantic cues are sparse. For instance, for {\gemini}~\citep{gemini}, accuracy drops from  \textcolor{teal}{$92.67\%$} on photos to \textcolor{red}{$76.49\%$} on symbolic sketches for the rotation task, and from \textcolor{teal}{$99.81\%$} to \textcolor{red}{$82.56\%$} for the scale task (Fig.~\ref{fig:main}).
    
    \item Among the transformations studied, rotation is consistently the most challenging, with models exhibiting particularly low performance even when performance on identity and scale tasks is much stronger.
    
    \item The observed failures persist across model architectures, model capacities, and prompting strategies, suggesting that the limitation is fundamental rather than a consequence of model size or prompt design.
\end{itemize}

\section{Related Work}
\label{sec:rel}

\noindent\textbf{Visual Reasoning in VLMs.}
The rapid advancement of vision-language models (VLMs) has prompted the question of whether these models exhibit true visual reasoning capabilities or primarily rely on learned statistical patterns. Prior work has explored the shortcomings of VLMs on seemingly simple visual reasoning tasks, including object counting and recognition \citep{bindingproblem, response_wide}. These failures have been further studied using tasks grounded in cognitive research \citep{doesspatialcognitionemerge, bindingproblem, wust2024bongard}, as well as benchmarks inspired by development psychology \citep{babyvision,yiu2024kiva}. 
However, these studies primarily evaluate visual tasks in semantically rich settings (e.g., real world photos)~\citep{Rotbench} and do not explicitly isolate if failure arises from limitations in visual reasoning or from reliance on semantic cues. 

\noindent\textbf{Are VLMs Transformation Invariant?}
Unlike Convolutional Neural Networks, which exhibit translational equivariance due to their architecture \citep{cnnequivariance}, vision transformers (ViTs) — the visual backbones of VLMs — do not have inherent transformation equivariance \citep{vitshiftequivariance}. Prior work suggests that ViTs may exhibit emergent invariance properties, particularly for rotation \citep{mentalrotation}. However, such invariance does not necessarily translate to robust tranformation reasoning, and VLMs fail at tasks that require visual invariance across a variety of settings \citep{babyvision, yiu2024kiva, doesspatialcognitionemerge, Rotbench}. While such failures have been documented in prior benchmarks, we instead evaluate how VLMs perform on transformation tasks across varying levels of semantic richness, revealing where these failures persist. Crucially, prior work~\citep{Rotbench, telling_left_right} do not study why VLMs fail at geometric reasoning nor if failure is uniform across visual domains.

\noindent\textbf{Evaluating Bias Across Visual Domains.}
Prior work on visual reasoning in VLMs has primarily focused on natural images~\citep{hemmat2024hidden, response_wide, Rotbench, telling_left_right} and does not systematically evaluate performance across diverse visual domains (e.g., photos, cartoons, sketches). When analyses have been performed across multiple domains, they typically focus on robustness to distribution shifts or stylistic variations \citep{mentalrotation, pixelspatternspoetry, visionlanguagemodelsbiased}, rather than isolating the role of semantic richness in shaping model behavior. 
In contrast, we evaluate VLM performance across a spectrum of semantic richness~\citep{PACS}, which allows us to disentangle geometric reasoning from reliance on semantic cues and exposes failure modes that remain hidden when evaluation is restricted to natural images.

\section{Studying VLM's invariance equivariance dilemma}
\label{sec:exp}
Given an image $I$, let $T$ denote a set of transformation functions that map an image to a transformed version $I'$. In our study, $T = \{t_{\text{rotation}}, t_{\text{scale}}, t_{\text{identity}}\}$, and $I' = t(I)$, where $t \in T$. 
Transformation equivariance refers to a model's ability to recognize that $I$ and $I'$ depict the same underlying content while also identifying the applied transformation $t \in T$. Since VLMs are trained on large-scale datasets of natural images~\citep{liu2023llava}, we hypothesize that their equivariance behavior may be influenced by learned data priors rather than consistent geometric reasoning. To evaluate this, we consider images with varying levels of semantic richness~\citep{PACS}, ranging from sparse sketches to photographs and abstract art. Below, we describe our experimental framework and corresponding research questions for evaluating these capabilities across different models.
\subsection{Experimental setup} \label{sec:setup}
\subsubsection{Datasets}\label{sec:datasets}
 We conduct our experiments on datasets of diverse semantic richness detailed below. 
\begin{itemize}
\itemsep-0.1em 
    \item \textbf{Omniglot}~\citep{omniglot} consists of handwritten characters from $50$ diverse scripts, ranging from widely recognized scripts such as Latin and Greek, to rarer scripts such as Manipuri and Glagolitic. It comprises $1,623$ distinct character classes in total, each containing multiple exemplar images rendered as black binary strokes on a uniform white background. We study Omniglot as a primary benchmark for two strategic reasons: \textbf{(a) Decouple spatial reasoning and script familiarity:} The dataset's inclusion of both more prevalent (e.g., Greek) and rare (e.g., Tagalog) scripts allows us to study whether VLM performance is driven by true geometric reasoning or merely by data familiarity. \textbf{ (b) Controlled Visual Stimuli:} Omniglot data is devoid of background noise, textures, and complex scenes, offering control over confounding variables and allowing us to study spatial reasoning in isolation.
       
\item \textbf{Times New Roman}~\citep{timesnewroman} is a dataset of English alphabet characters rendered in the standard Times New Roman typeface, thus offering a single, consistent canonical appearance. This dataset: (a) mirrors the \textbf{extreme prevalence} of English typography in web-scale pretraining data, and (b) the \textbf{structural precision} of digital fonts removes stroke ambiguity and serves as a controlled baseline to isolate the model's geometric reasoning capabilities.

\item\textbf{Handwritten English}~\citep{handwritten_english_characters_digits} characters and digits complement Omniglot. Unlike Times New Roman, this dataset exhibits stroke-level variation characteristic of handwriting, while retaining the familiarity of English characters and digits.

\item\textbf{PACS}~\citep{PACS} dataset comprises $9,991$ images of seven object categories (e.g., dog, guitar, elephant) spanning four visual domains: Photograph, Art Painting, Cartoon, and Sketch. Unlike the above character datasets, PACS offers high-level semantic content and exhibits substantial variation in texture, color, and level of visual abstraction. By evaluating models on this dataset, we can systematically analyze how geometric reasoning fluctuates for a single object category (e.g., dog) as its semantic representation becomes increasingly rich and complex.
\end{itemize}

Due to the high API costs of closed-source models, we sample a subset of each dataset for evaluation. For Omniglot, we sample one handwritten example from each character class across all $50$ scripts, thereby evaluating on a total of $1,623$ characters. For Handwritten English, we sample one handwritten exemplar per character, yielding  $52$ characters in total (both lower and upper-case characters). For Times New Roman, we evaluate on all $52$ canonical characters from the English alphabet. For PACS~\citep{PACS} we randomly sample $200$ images from each of the four domains, maintaining a balanced split across all $7$ object categories within each domain, resulting in $800$ images in total.
We release these evaluation splits as \textbf{SERGE} (\textbf{SE}mantic \textbf{R}ichness \& \textbf{G}eometric \textbf{E}valuation.)\footnote{%
  \textbf{SE}mantic \textbf{R}ichness \& \textbf{G}eometric \textbf{E}valuation.
  Available at \url{https://huggingface.co/datasets/XThomasBU/SERGE}.%
}, our benchmark for probing visual invariance across the semantic-richness spectrum.

\noindent\textbf{Evaluation setup.} Each evaluation instance consists of a pair of images $(I, I')$ and a text prompt (described later). The first image $I$ is sampled from one of the above datasets. The second image $I'$ is either (a) a \textbf{positive sample}, $t(I)$ for $t \in T$, or (b) a \textbf{negative sample}, $t(J)$, a transformed version of a different image $J$. We prompt each model to determine if the two images depict the same underlying character or object under transformation $t$. For Handwritten English and Times New Roman, we avoid constructing negative pairs from upper and lower case variants of the same letter (e.g., `a' and `A'), as they correspond to the same underlying character and would make negatives ambiguous. A true positive (TP) corresponds to correctly identifying $(I, t(I))$ as the same character or object, while a true negative (TN) corresponds to correctly identifying $(I, t(J))$ as different. False positives (FP) and false negatives (FN) denote the respective misclassifications. We evaluate performance using the following metrics:
\vspace{-0.1in}
\begin{itemize}
\itemsep-0.1em 
    \item \textbf{Recall (TPR).} $\text{TPR} = \text{TP} / (\text{TP} + \text{FN})$.
    \item \textbf{Specificity (TNR).} $\text{TNR} = \text{TN} / (\text{TN} + \text{FP})$.
    \item \textbf{Accuracy.} $\text{Accuracy} = (\text{TP} + \text{TN}) / (\text{TP} + \text{TN} + \text{FP} + \text{FN})$.
\end{itemize}
\vspace{-0.1in}
\noindent \textbf{Models studied:}  We study two powerful closed-sourced models: \gemini~\citep{gemini} and \gpt~\citep{gpt5}, and two open-sourced models \qwenTwo~\citep{qwen2}, and \qwenThree~\citep{qwen3}. Within the Qwen Series, we study different variants based on the number of parameters: \qwenSeven, \qwenThirtyTwo,  \qwenThreeEight, \qwenThreeThirty. Studying diverse models helps us understand the pervasiveness of current failures at various model capacities. 

\subsection{Studying transformation invariance in MLLMs}\label{sec:trans_inv}
We use \textbf{rotation} as our main transformation to analyze the ability of Multi-Modal Large Language Models (MLLMs) to perceive transformations. Specifically,  given two images, the task is to identify if the second one could be a rotated version of the first image. We study other transformations in Sec.~\ref{sec:other_transform}.

\begin{figure}[!t]
\vspace{-1.2em}
  \centering
  \includegraphics[width=0.99\linewidth, keepaspectratio]{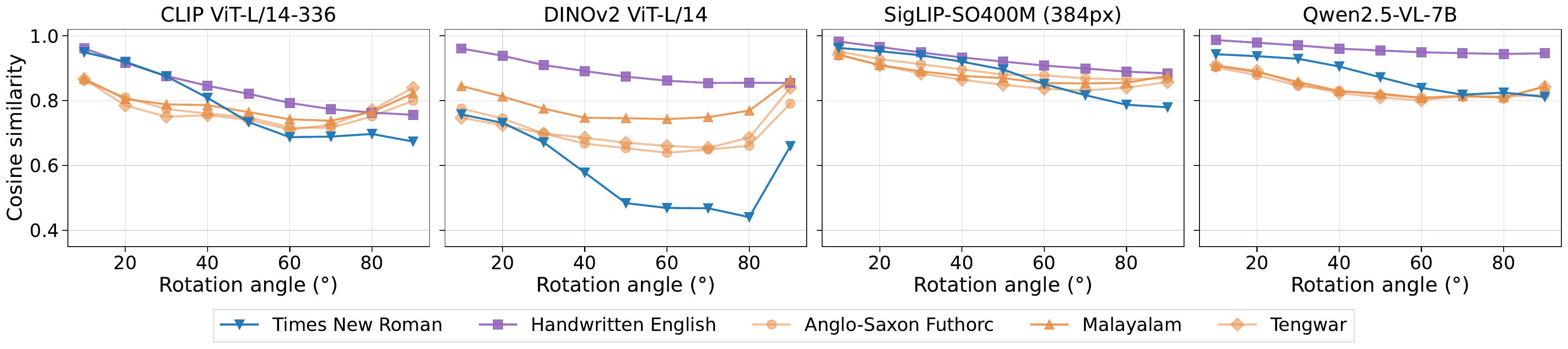}
\vspace{-1.0em}
\caption{\textbf{Cosine similarity between features extracted from different vision encoders} on pairs of images under rotation. Select \textcolor{scriptorange}{Omniglot scripts} are shown in orange, while \textcolor{scriptblue}{Times New Roman} and \textcolor{scriptpurple}{Handwritten English} are shown in blue and purple respectively. Across all encoders, similarity decreases with increasing rotation angle, with DINOv2 showing the steepest drop and SigLIP and {\qwenSeven} maintaining relatively higher similarity.}
\vspace{+0.4em}
\label{fig:encoder_graphs_omniglot}
\end{figure}

\subsubsection{Are vision encoders rotational invariant?}\label{sec:enc_rot}
To begin, we study how vision components within MLLMs behave under rotation. We compare vision encoders including {\clipL}~\citep{clip}, {\dino}~\citep{dino}, {\siglip}~\citep{siglip}, and the visual encoder of {\qwenSeven}~\citep{qwen2} on the Omniglot dataset. For each encoder, we pass images through the pretrained backbone and extract a global image representation. We use the \texttt{[CLS]} token for CLIP and DINOv2, the multi-head attention pooling (MAP) output
for SigLIP, and the mean of all image token features for {\qwenSeven}. 

\noindent \textbf{Findings:} 
From Fig.~\ref{fig:encoder_graphs_omniglot}, vision encoders exhibit high cosine similarity at smaller rotation angles ($0\degree$--$30\degree$), indicating that representations remain similar under minor transformations. As the rotation angle increases, similarity decreases across all encoders. DINOv2 shows the largest drop, particularly for Times New Roman, whereas SigLIP and {\qwenSeven} maintain relatively higher similarity at larger angles. English character sets (Times New Roman and Handwritten English) exhibit a more consistent monotonic decline with increasing rotation, while Omniglot scripts show non-monotonic trends. 
We next examine whether these feature similarities under rotation are sufficient for accurate transformation reasoning once coupled with a language decoder in MLLMs.

\subsubsection{Are vision language models (VLMs) rotational invariant?}
\label{sec:rot}

\begin{table}[!t]
\vspace{-1.0em}
\centering
\LARGE
\resizebox{\columnwidth}{!}{
\begin{tabular}{lcccccccccccccccccc}
\toprule
& \multicolumn{3}{>{\columncolor{cyan!10}}c}{\textbf{\qwenSeven}} 
& \multicolumn{3}{>{\columncolor{cyan!10}}c}{\textbf{\qwenThirtyTwo}}
& \multicolumn{3}{>{\columncolor{blue!10}}c}{\textbf{\qwenThreeEight}}
& \multicolumn{3}{>{\columncolor{blue!10}}c}{\textbf{\qwenThreeThirty}}
& \multicolumn{3}{>{\columncolor{gray!10}}c}{\textbf{\gpt}}
& \multicolumn{3}{>{\columncolor{gray!10}}c}{\textbf{\gemini}} \\
\cmidrule(lr){2-4} 
\cmidrule(lr){5-7}
\cmidrule(lr){8-10}
\cmidrule(lr){11-13}
\cmidrule(lr){14-16}
\cmidrule(lr){17-19}

\textbf{Dataset} 
& \textbf{Acc.} & \textbf{TNR} & \textbf{TPR}
& \textbf{Acc.} & \textbf{TNR} & \textbf{TPR}
& \textbf{Acc.} & \textbf{TNR} & \textbf{TPR}
& \textbf{Acc.} & \textbf{TNR} & \textbf{TPR}
& \textbf{Acc.} & \textbf{TNR} & \textbf{TPR}
& \textbf{Acc.} & \textbf{TNR} & \textbf{TPR} \\
\midrule

Times New Roman
& \textcolor{teal}{51.07} & 100.00 & 02.14
& \textcolor{red}{52.67} & 100.00 & 05.34
& \textcolor{red}{50.11} & 100.00 & 00.21
& \textcolor{teal}{65.81} & 100.00 & 31.62
& \textcolor{red}{74.25} & 95.09 & 53.42
& \textcolor{teal}{89.32} & 100.00 & 78.63 \\

Handwritten English
& 50.85 & 98.08 & 03.63
& \textcolor{teal}{62.50} & 98.08 & 26.92
& \textcolor{red}{50.00} & 100.00 & 00.00
& 55.98 & 100.00 & 11.97
& 67.84 & 96.58 & 39.10
& 68.27 & 99.57 & 36.97 \\

Omniglot
& \textcolor{red}{50.72} & 99.46 & 01.98
& 54.17 & 94.71 & 13.62
& \textcolor{teal}{51.01} & 99.96 & 02.06
& \textcolor{red}{56.64} & 99.75 & 13.53
& \textcolor{teal}{75.55} & 80.91 & 70.19
& \textcolor{red}{76.90} & 98.69 & 55.10 \\

\noalign{\vskip 3pt}
\cdashline{1-19}
\noalign{\vskip 2pt}

Random guess 
& 50.00 & 50.00 & 50.00
& 50.00 & 50.00 & 50.00
& 50.00 & 50.00 & 50.00
& 50.00 & 50.00 & 50.00
& 50.00 & 50.00 & 50.00
& 50.00 & 50.00 & 50.00 \\

\bottomrule
\end{tabular}
}

\vspace{-0.4em}
\caption{
\textbf{Rotation recognition performance across character datasets.} Best and worst accuracy per model across datasets are highlighted in \textcolor{teal}{teal} and \textcolor{red}{red}, respectively. Performance is aggregated over rotation angles $10^\circ$–$90^\circ$. While TNR remains near-perfect across all models, TPR is consistently low, indicating a failure to recognize rotated variants. Closed-source models perform better than open-source models, but the failure persists across all models. }
\vspace{-0.5em}
\label{tab:rotation_results}
\end{table}

We use the data described in Sec.~\ref{sec:setup} to evaluate rotational invariance on 50 handwritten scripts from the Omniglot~\citep{omniglot} dataset, along with the Handwritten English alphabet and the Times New Roman printed alphabet. We test on both positive and negative cases -- i.e., where the two images are rotated versions of one another (ground truth response: ``yes'') or where they are two different characters (ground truth response: ``no''). 
We use the below prompt for the rotation invariance task, and test for $10^\circ$–$90^\circ$ rotations.
\begin{tcolorbox}[
    colback=gray!8,
    colframe=gray!50,
    boxrule=1pt,
    arc=3pt,
    left=7pt,
    right=7pt,
    top=4pt,
    bottom=4pt,
]
\small 
\textbf{Rotation Recognition Prompt}
\textit{If I rotate the first image, can I get the second image? Answer in curly brackets, e.g. \{Yes\} or \{No\}.} 
\end{tcolorbox}
\label{sec:Omniglotprompt}

\noindent\textbf{Findings:} 
Table~\ref{tab:rotation_results} reports accuracy, TNR, and TPR across all datasets, with $50.00\%$ corresponding to random guessing. Across all models, TNR remains high, while TPR is consistently low, indicating a strong bias toward predicting ``\{No\}''. This results in near-random accuracy despite high TNR. For instance, {\qwenThirtyTwo} achieves a TPR of $5.34\%$ on Times New Roman and $13.62\%$ on Omniglot, while {\qwenThreeThirty} reaches only $31.62\%$ and $13.53\%$, respectively.
Although closed-source models perform substantially better, even these models fail to achieve consistently high TPR across datasets: {\gpt} achieves a TPR of $53.42\%$ on Times New Roman and $70.19\%$ on Omniglot, while {\gemini} reaches $78.63\%$ and $55.10\%$, respectively. 
Overall, these results indicate that identifying rotated variants of the same character remains a challenging task for current VLMs, reflecting a lack of robust geometric reasoning. 

\begin{table}[!t]
\centering
\LARGE

\resizebox{\textwidth}{!}{%
\begin{tabular}{lcccccccccccccccccc}
\toprule
& \multicolumn{3}{>{\columncolor{cyan!10}}c}{\textbf{\qwenSeven}} 
& \multicolumn{3}{>{\columncolor{cyan!10}}c}{\textbf{\qwenThirtyTwo}}
& \multicolumn{3}{>{\columncolor{blue!10}}c}{\textbf{\qwenThreeEight}}
& \multicolumn{3}{>{\columncolor{blue!10}}c}{\textbf{\qwenThreeThirty}}
& \multicolumn{3}{>{\columncolor{gray!10}}c}{\textbf{\gpt}}
& \multicolumn{3}{>{\columncolor{gray!10}}c}{\textbf{\gemini}} \\
\cmidrule(lr){2-4}
\cmidrule(lr){5-7}
\cmidrule(lr){8-10}
\cmidrule(lr){11-13}
\cmidrule(lr){14-16}
\cmidrule(lr){17-19}

\textbf{Domain} 
& \textbf{Acc.} & \textbf{TNR} & \textbf{TPR}
& \textbf{Acc.} & \textbf{TNR} & \textbf{TPR}
& \textbf{Acc.} & \textbf{TNR} & \textbf{TPR}
& \textbf{Acc.} & \textbf{TNR} & \textbf{TPR}
& \textbf{Acc.} & \textbf{TNR} & \textbf{TPR}
& \textbf{Acc.} & \textbf{TNR} & \textbf{TPR} \\
\midrule

Photo
& \textcolor{teal}{59.42} & 100.00 & 18.83
& \textcolor{teal}{77.67} & 100.00 & 55.33
& \textcolor{teal}{50.25} & 100.00 & 00.50
& \textcolor{teal}{64.58} & 100.00 & 29.17
& \textcolor{teal}{99.50} & 100.00 & 99.00
& 92.67 & 100.00 & 85.33 \\

Art Painting
& \textcolor{red}{50.25} & 100.00 & 00.50
& 63.50 & 100.00 & 27.00
& \textcolor{red}{50.08} & 100.00 & 00.17
& 55.08 & 100.00 & 10.17
& \textcolor{teal}{99.50} & 100.00 & 99.00
& \textcolor{teal}{94.17} & 100.00 & 88.33 \\

Cartoon
& 51.42 & 100.00 & 02.83
& 70.08 & 100.00 & 40.17
& \textcolor{teal}{50.25} & 100.00 & 00.50
& 60.67 & 100.00 & 21.33
& 98.17 & 100.00 & 96.33
& 90.33 & 100.00 & 80.67 \\

Sketch
& 52.25 & 100.00 & 04.50
& \textcolor{red}{57.25} & 100.00 & 14.50
& 50.42 & 100.00 & 00.83
& \textcolor{red}{52.83} & 100.00 & 05.67
& \textcolor{red}{92.25} & 99.67 & 84.83
& \textcolor{red}{86.50} & 99.83 & 73.17 \\

\noalign{\vskip 3pt}
\cdashline{1-19}
\noalign{\vskip 2pt}

Random guess 
& 50.00 & 50.00 & 50.00
& 50.00 & 50.00 & 50.00
& 50.00 & 50.00 & 50.00
& 50.00 & 50.00 & 50.00
& 50.00 & 50.00 & 50.00
& 50.00 & 50.00 & 50.00 \\

\bottomrule
\end{tabular}}

\vspace{-0.4em}
\caption{\textbf{Rotation recognition performance across PACS domains.} Performance aggregated over rotation angles $90^\circ$, $180^\circ$, and $270^\circ$. While TNR remains near-perfect across models, TPR varies significantly across domains, with strong performance on photos and substantial degradation on sketches, indicating reliance on semantic cues rather than true geometric reasoning.}
\vspace{-0.8em}
\label{tab:pacs_results}
\end{table}

\subsubsection{Suspect 1: Data bias} 
Models are trained primarily on {natural images} containing objects, complex scene structure, and textures~\citep{liu2023llava}. This raises the question: could it be that rotational invariance is domain-specific? To disentangle this, we replicate the study in Sec.~\ref{sec:rot} on PACS~\citep{PACS} using three rotations: $90\degree$, $180\degree$, and $270\degree$ across different visual domains (Fig.~\ref{fig:data_grid}). These angles avoid \textit{blank} padding pixels introduced by non-orthogonal rotations (e.g., $45\degree$) (see suppl.). While this was not an issue for Omniglot~\citep{omniglot} due to its uniform white background, such artifacts could otherwise confound performance on PACS images.

\textbf{Findings:} From Table~\ref{tab:pacs_results}, we observe almost no variation in TNR across models and domains, while TPR and accuracy vary substantially. Larger {\qwenTwo} and {\qwenThree} models improve over their smaller counterparts, but remain significantly behind closed-source models such as {\gpt} and {\gemini}. 
Across all models, performance is highest on photos, followed by cartoons and art paintings, and consistently lowest on sketches. For example, {\qwenThirtyTwo} achieves a TPR of $55.33\%$ on photos but only $14.50\%$ on sketches, while {\qwenThreeThirty} drops from $29.17\%$ to $5.67\%$ from photos to sketches. Even for closed-source models, this gap persists: {\gpt} achieves $99.00\%$ TPR on photos but drops to $84.83\%$ on sketches, and {\gemini} drops from $85.33\%$ to $73.17\%$, respectively.
Notably, closed-source models also exhibit a large drop when evaluated on symbolic sketches compared to semantic sketches. For instance, {\gpt} achieves a TPR of $84.83\%$ on PACS sketches, but only $70.19\%$ on Omniglot scripts (Table~\ref{tab:rotation_results}), indicating that performance degrades further as semantic richness decreases. This trend suggests that even strong models rely on semantic cues and data familiarity, and struggle when forced to operate purely on geometric structure.

\subsubsection{Suspect 2: Model capacity}
We study different variants from the family of {\qwenTwo} and {\qwenThree} models and report the effect of model capacity on rotational invariance. 

\noindent \textbf{Findings:} From Table~\ref{tab:rotation_results}, we observe a modest increase in accuracy with model capacity. For example, on Handwritten English, accuracy improves from $50.85\%$ ({\qwenSeven}) to $62.50\%$ ({\qwenThirtyTwo}), and on Times New Roman from $50.11\%$ ({\qwenThreeEight}) to $65.81\%$ ({\qwenThreeThirty}). Similar trends are observed across datasets. However, TPR remains consistently low. 
Additionally, as reported in Table~\ref{tab:pacs_results}, while TPR is higher for {\qwenThirtyTwo} and {\qwenThreeThirty} on real photos, with values of $55.33\%$ and $29.17\%$, respectively, performance on sketches remains poor at $14.50\%$ and $5.67\%$, indicating that scale alone is insufficient to improve robust geometric reasoning in VLMs.



\subsubsection{Suspect 3: Brittleness to prompt}\label{sec:brittle}
Next, we study whether the performance discrepancy between VLMs and visual encoders stems from \textit{how} the task is phrased. Specifically, we wish to disentangle if a model's understanding of rotation transformation is tied to its object recognition capability. On PACS dataset~\citep{PACS} (Sec.~\ref{sec:datasets}), we use the following prompts:

\begin{tcolorbox}[
    colback=gray!8,
    colframe=gray!50,
    boxrule=0.5pt,
    arc=2pt,
    left=6pt,
    right=6pt,
    top=4pt,
    bottom=4pt,
]
\small
\centering
\textbf{Object Recognition focused (OR):} \label{sec:PACSprompts}
\textit{Identify the main object in these images. Return ONLY the object name. Your options are: Dog, Elephant, Giraffe, Horse, Person, Guitar, or House.} 
\\
\vspace{0.5em}
\textbf{Object Identification (OI):} 
\textit{Is the object in these images a {\{class\_name\}}? Answer ONLY `Yes' or `No'.} 
\\
\vspace{0.5em}

\textbf{Rotation Recognition (RR):}
\textit{If I rotate the first image, can I get the second image? Answer in curly brackets, e.g. \{Yes\} or \{No\}.}
\end{tcolorbox}

OR is formulated as a multi-class classification task without explicit negative samples, whereas OI and RR are binary tasks with both positive and negative samples (Sec.~\ref{sec:setup}). The object categories used in the OR and OI prompts are from the PACS dataset. We hypothesize that object identification is a simpler task than recognition, enabling a more controlled assessment of VLM capabilities. Across all three tasks, we use images $(I, I')$ as described for the rotation task ($t_{\text{rotation}}$) in Sec.~\ref{sec:datasets}.

\begin{table}[!t]
\vspace{-1.2em}
\centering
\LARGE
\resizebox{\columnwidth}{!}{
\begin{tabular}{lcccccccccccccccccc}
\toprule
& \multicolumn{3}{>{\columncolor{cyan!10}}c}{\textbf{\qwenSeven}} 
& \multicolumn{3}{>{\columncolor{cyan!10}}c}{\textbf{\qwenThirtyTwo}}
& \multicolumn{3}{>{\columncolor{blue!10}}c}{\textbf{\qwenThreeEight}}
& \multicolumn{3}{>{\columncolor{blue!10}}c}{\textbf{\qwenThreeThirty}}
& \multicolumn{3}{>{\columncolor{gray!10}}c}{\textbf{\gpt}}
& \multicolumn{3}{>{\columncolor{gray!10}}c}{\textbf{\gemini}} \\
\cmidrule(lr){2-4}
\cmidrule(lr){5-7}
\cmidrule(lr){8-10}
\cmidrule(lr){11-13}
\cmidrule(lr){14-16}
\cmidrule(lr){17-19}

\textbf{Domain} 
& \textbf{OR} & \textbf{RR} & \textbf{OI}
& \textbf{OR} & \textbf{RR} & \textbf{OI}
& \textbf{OR} & \textbf{RR} & \textbf{OI}
& \textbf{OR} & \textbf{RR} & \textbf{OI}
& \textbf{OR} & \textbf{RR} & \textbf{OI}
& \textbf{OR} & \textbf{RR} & \textbf{OI} \\
\midrule

Photo
& \textcolor{teal}{100.00} & \textcolor{red}{59.42} & 98.75
& \textcolor{teal}{100.00} & \textcolor{red}{77.67} & 99.75 
& \textcolor{teal}{100.00} & \textcolor{red}{50.25} & 99.25
& \textcolor{teal}{100.00} & \textcolor{red}{64.58} & 99.00
& \textcolor{teal}{100.00} & \textcolor{red}{99.50} & 99.25
& \textcolor{teal}{100.00} & \textcolor{red}{92.67} & 98.25 \\

Art Painting
& \textcolor{teal}{98.00} & \textcolor{red}{50.25} & 96.75
& \textcolor{teal}{99.50} & \textcolor{red}{63.50} & 97.50
& \textcolor{teal}{99.50} & \textcolor{red}{50.08} & 98.00
& \textcolor{teal}{98.50} & \textcolor{red}{55.08} & 98.50
& 98.50 & \textcolor{teal}{99.50} & \textcolor{red}{97.50}
& \textcolor{teal}{100.00} & \textcolor{red}{94.17} & 97.00 \\

Cartoon
& \textcolor{teal}{99.50} & \textcolor{red}{51.42} & 97.00
& \textcolor{teal}{99.00} & \textcolor{red}{70.08} & 98.00
& \textcolor{teal}{100.00} & \textcolor{red}{50.25} & 97.75
& \textcolor{teal}{100.00} & \textcolor{red}{60.67} & 97.75
& \textcolor{teal}{100.00} & 98.17 & \textcolor{red}{97.75} 
& \textcolor{teal}{100.00} & \textcolor{red}{90.33} & 98.25 \\

Sketch
& \textcolor{teal}{94.50} & \textcolor{red}{52.25} & 92.75
& \textcolor{teal}{96.50} & \textcolor{red}{57.25} & 86.25
& \textcolor{teal}{96.50} & \textcolor{red}{50.42} & 87.00
& \textcolor{teal}{96.00} & \textcolor{red}{52.83} & 91.00
& 94.50 & \textcolor{red}{92.25} & \textcolor{teal}{95.50}
& \textcolor{teal}{97.50} & \textcolor{red}{86.50} & 96.00 \\

\noalign{\vskip 3pt}
\cdashline{1-19}
\noalign{\vskip 2pt}

Random guess 
& 50.00 & 50.00 & 50.00
& 50.00 & 50.00 & 50.00
& 50.00 & 50.00 & 50.00
& 50.00 & 50.00 & 50.00
& 50.00 & 50.00 & 50.00
& 50.00 & 50.00 & 50.00 \\

\bottomrule
\end{tabular}
}

\vspace{-0.4em}
\caption{\textbf{Performance across task formulations on PACS.} Accuracy is reported for object recognition (OR), object identification (OI), and rotation recognition (RR) tasks (Sec.~\ref{sec:brittle}) across PACS domains. For each model and domain, the best and worst task performances are highlighted in \textcolor{teal}{teal} and \textcolor{red}{red}, respectively. While models perform near-perfectly on OR and OI, performance drops significantly on RR, highlighting a gap between semantic recognition and geometric reasoning.}
\label{tab:brittleness_pacs_results}
\end{table}
\noindent \textbf{Findings:} 
From Table~\ref{tab:brittleness_pacs_results}, we note that all models achieve near-perfect performance on object recognition (OR) and object identification (OI) across domains, whereas rotation recognition (RR) performance is substantially lower. For instance, in the Photo domain, RR accuracy is $59.42\%$ and $77.67\%$ for {\qwenSeven} and {\qwenThirtyTwo}, but drops to $50.25\%$ for {\qwenThreeEight}, compared to near-perfect performance on OR and OI. This trend persists across all domains, with RR consistently lower than OR and OI, with the exception of {\gpt}. The finding is clear: models achieve near-perfect accuracy on object recognition (OR) and identification (OI) tasks while performance collapses on rotation recognition (RR) where such explicit cues are absent. This divergence suggests that models rely on semantic recognition as a shortcut rather than genuine transformation reasoning.

\begin{figure}[!h]
\vspace{-2mm}
\centering
\noindent\textbf{Takeaways from Studying Rotational Invariance of VLMs}
\begin{tcolorbox}[
    colframe=teal!60!black,
colback=teal!8,
    arc=6pt,              
    boxrule=0.8pt,
    width=\columnwidth,
    left=4pt, right=4pt, top=4pt, bottom=4pt
]
\begin{enumerate}[leftmargin=1.5em]
\setlength\itemsep{0.45em}
\footnotesize
    \item VLMs fail ungracefully when presented with images which are rotated variants of each other across datasets, model capacities, and prompts.
    \item All models, across different capacities, perform well on real-world photos and the worst on symbolic sketches, indicating lack of true geometric understanding. 
    \item The underlying visual encoder of {\qwenSeven} appears to retain the invariance, but this signal is lost when paired with language decoder.
    \item Models rely on semantic recognition as a shortcut rather than true transformation reasoning.
\end{enumerate}
\end{tcolorbox}
\vspace{-0.25in}
\end{figure}

\subsection{Is this behavior specific to rotation transformation?} 
\label{sec:other_transform}

\subsubsection{Case Study 1: Identity Transformation} \label{sec:identity_exp}
\begin{figure}[!t]
\vspace{-1.2em}
  \centering
  \includegraphics[width=0.55\linewidth, keepaspectratio]{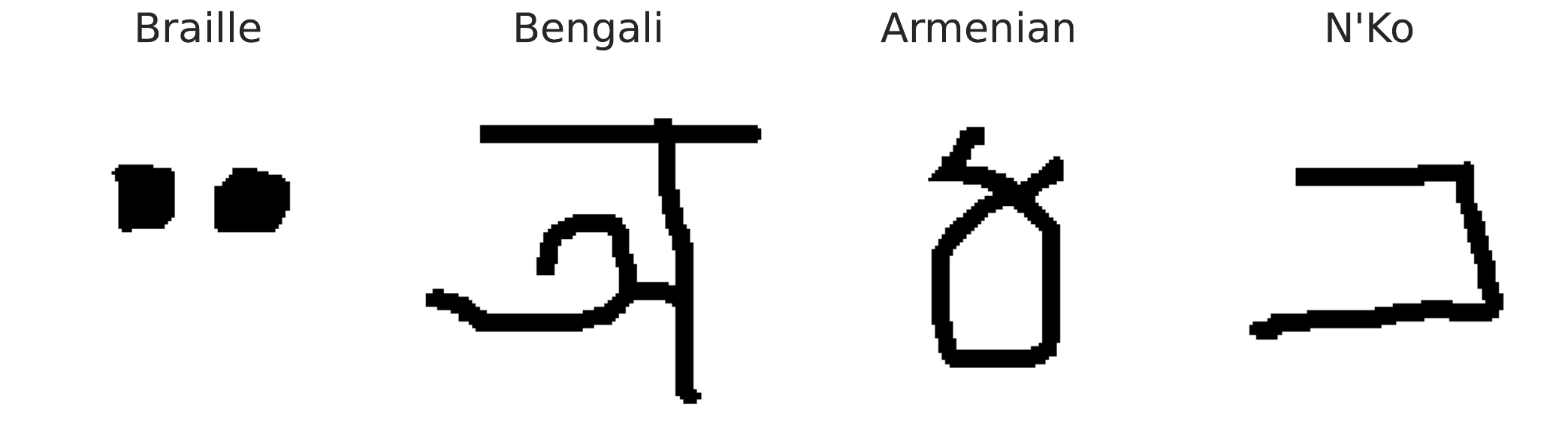}
\vspace{-1.4em}
\caption{\textbf{Failure cases on the identity task (Sec.~\ref{sec:identity_exp}) for {\qwenSeven}.} We show four randomly selected examples from Omniglot dataset where the model incorrectly predicts that two identical inputs correspond to different characters.}
\vspace{-0.4em}
  \label{fig:failure}
\end{figure}

We begin with a basic visual task: given \textbf{two identical images} of the same character or object, can present day VLMs identify that they are the same? We study this question using the Omniglot, Times New Roman, and Handwritten English datasets listed in Sec.~\ref{sec:datasets}, using the following prompt 
:
\begin{tcolorbox}[
    colback=gray!8,
    colframe=gray!50,
    boxrule=0.5pt,
    arc=2pt,
    left=6pt,
    right=6pt,
    top=4pt,
    bottom=4pt,
]
\small
\centering
\textbf{Identity Prompt:}
\textit{Are these the same characters? Answer in curly brackets, e.g. \{Yes\} or \{No\}.}
\end{tcolorbox}

\begin{table}[!t]
\centering
\LARGE

\resizebox{0.99\textwidth}{!}{%
\begin{tabular}{lcccccccccccccccccc}
\toprule
& \multicolumn{3}{>{\columncolor{cyan!10}}c}{\textbf{\qwenSeven}} 
& \multicolumn{3}{>{\columncolor{cyan!10}}c}{\textbf{\qwenThirtyTwo}}
& \multicolumn{3}{>{\columncolor{blue!10}}c}{\textbf{\qwenThreeEight}}
& \multicolumn{3}{>{\columncolor{blue!10}}c}{\textbf{\qwenThreeThirty}}
& \multicolumn{3}{>{\columncolor{gray!10}}c}{\textbf{\gpt}}
& \multicolumn{3}{>{\columncolor{gray!10}}c}{\textbf{\gemini}} \\
\cmidrule(lr){2-4} 
\cmidrule(lr){5-7}
\cmidrule(lr){8-10}
\cmidrule(lr){11-13}
\cmidrule(lr){14-16}
\cmidrule(lr){17-19}

\textbf{Dataset} 
& \textbf{Acc.} & \textbf{TNR} & \textbf{TPR}
& \textbf{Acc.} & \textbf{TNR} & \textbf{TPR}
& \textbf{Acc.} & \textbf{TNR} & \textbf{TPR}
& \textbf{Acc.} & \textbf{TNR} & \textbf{TPR}
& \textbf{Acc.} & \textbf{TNR} & \textbf{TPR}
& \textbf{Acc.} & \textbf{TNR} & \textbf{TPR} \\
\midrule

Times New Roman
& \textcolor{teal}{100.00} & 100.00 & 100.00
& \textcolor{teal}{100.00} & 100.00 & 100.00
& \textcolor{teal}{100.00} & 100.00 & 100.00
& \textcolor{teal}{100.00} & 100.00 & 100.00
& \textcolor{teal}{100.00} & 100.00 & 100.00
& \textcolor{teal}{100.00} & 100.00 & 100.00 \\

Handwritten English
& 99.04 & 98.08 & 100.00
& 99.04 & 98.08 & 100.00
& \textcolor{teal}{100.00} & 100.00 & 100.00
& 99.04 & 98.08 & 100.00
& 99.04 & 98.08 & 100.00
& 99.04 & 98.08 & 100.00 \\

Omniglot
& \textcolor{red}{81.33} & 99.88 & 62.78
& \textcolor{red}{96.95} & 94.52 & 99.38
& 99.72 & 99.45 & 100.00
& 99.57 & 99.88 & 99.26
& \textcolor{red}{98.77} & 97.54 & 100.00
& \textcolor{red}{96.67} & 93.35 & 100.00 \\

\noalign{\vskip 3pt}
\cdashline{1-19}
\noalign{\vskip 2pt}

Random guess 
& 50.00 & 50.00 & 50.00
& 50.00 & 50.00 & 50.00
& 50.00 & 50.00 & 50.00
& 50.00 & 50.00 & 50.00
& 50.00 & 50.00 & 50.00
& 50.00 & 50.00 & 50.00 \\

\bottomrule
\end{tabular}}

\vspace{-0.4em}
\caption{\textbf{Model performance for the identity transformation on printed and handwritten scripts.} All models achieve perfect performance on the Times New Roman dataset. The models achieve near-perfect performance on the Handwritten English dataset with a few mistakes in the negative case. The worst performances across all models occur with the Omniglot dataset.}
\vspace{-0.4em}
\label{tab:identity_results}
\end{table}

\noindent\textbf{Results.} Table~\ref{tab:identity_results} reports model performance on the identity task. Times New Roman achieves $100\%$ accuracy across all six models, and Handwritten English follows closely with $99.04\%$. For Omniglot, all models achieve near-perfect performance with the exception of \qwenSeven, which records a TPR of $62.78\%$, indicating that it incorrectly rejects a substantial fraction of identical character pairs. We show $4$ random visual examples in Fig.~\ref{fig:failure} where {\qwenSeven} fails to identify that the two images are identical.
We believe that the performance collapse of {\qwenSeven} on Omniglot on a simple identity matching task indicates its lack of geometric grounding. It confirms that the model’s success is anchored in script-specific memorization rather than true geometric reasoning.

\subsubsection{Case Study 2: Scale Invariance}
\label{sec:scale_exp}
An ideal MLLM should recognize a character's identity invariant of scale, performing consistently across both familiar (Latin) and unfamiliar (Mongolian) scripts. We study this on Times New Roman, Handwritten English, and Omniglot~\citep{omniglot} (Sec.~\ref{sec:datasets}). Each sample pairs a full-resolution reference image ($1.0 \times$) with a query image containing a character—either identical or different—at a reduced scale $s \in \{0.1, 0.3, 0.5, 0.9\}$. Scaled characters are padded with white pixels to maintain original image dimensions (Fig.~\ref{fig:main}) (more details in suppl.). We use the following prompt:

\begin{tcolorbox}[
    colback=gray!8,
    colframe=gray!50,
    boxrule=0.5pt,
    arc=2pt,
    left=6pt,
    right=6pt,
    top=4pt,
    bottom=4pt,
]
\small
\textbf{Scale prompt.}
\textit{Compare the two images and decide if they show the same character. Ignore differences in scale, size, or resolution. Answer with exactly YES or NO.}
\end{tcolorbox}

\begin{table}[!t]
\centering
\LARGE
\resizebox{0.99\linewidth}{!}{

\begin{tabular}{lcccccccccccccccccc}
\toprule
& \multicolumn{3}{>{\columncolor{cyan!10}}c}{\textbf{\qwenSeven}} 
& \multicolumn{3}{>{\columncolor{cyan!10}}c}{\textbf{\qwenThirtyTwo}}
& \multicolumn{3}{>{\columncolor{blue!10}}c}{\textbf{\qwenThreeEight}}
& \multicolumn{3}{>{\columncolor{blue!10}}c}{\textbf{\qwenThreeThirty}}
& \multicolumn{3}{>{\columncolor{gray!10}}c}{\textbf{GPT-5.2}}
& \multicolumn{3}{>{\columncolor{gray!10}}c}{\textbf{\gemini}} \\
\cmidrule(lr){2-4}
\cmidrule(lr){5-7}
\cmidrule(lr){8-10}
\cmidrule(lr){11-13}
\cmidrule(lr){14-16}
\cmidrule(lr){17-19}

\textbf{Dataset}
& \textbf{Acc.} & \textbf{TNR} & \textbf{TPR}
& \textbf{Acc.} & \textbf{TNR} & \textbf{TPR}
& \textbf{Acc.} & \textbf{TNR} & \textbf{TPR}
& \textbf{Acc.} & \textbf{TNR} & \textbf{TPR}
& \textbf{Acc.} & \textbf{TNR} & \textbf{TPR}
& \textbf{Acc.} & \textbf{TNR} & \textbf{TPR} \\
\midrule

Times New Roman
& \textcolor{teal}{98.80} & 97.60 & 100.00
& \textcolor{teal}{99.76} & 99.52 & 100.00
& \textcolor{teal}{100.00} & 100.00 & 100.00
& \textcolor{teal}{98.79} & 97.59 & 100.00
& \textcolor{teal}{98.79} & 99.03 & 98.55
& \textcolor{teal}{99.51} & 100.00 & 99.03 \\

Handwritten English
& 96.77 & 95.56 & 97.98
& 95.36 & 92.34 & 98.39
& 97.59 & 95.19 & 100.00
& 97.59 & 99.03 & 96.15
& 98.07 & 98.07 & 98.07
& 96.63 & 98.07 & 95.19 \\

Omniglot
& \textcolor{red}{77.40} & 92.99 & 61.81
& \textcolor{red}{74.21} & 67.87 & 80.55
& \textcolor{red}{76.05} & 97.11 & 54.99
& \textcolor{red}{77.04} & 95.44 & 58.64
& \textcolor{red}{79.72} & 93.20 & 66.23
& \textcolor{red}{82.56} & 90.11 & 75.01 \\

\noalign{\vskip 3pt}
\cdashline{1-19}
\noalign{\vskip 2pt}

Random guess 
& 50.00 & 50.00 & 50.00
& 50.00 & 50.00 & 50.00
& 50.00 & 50.00 & 50.00
& 50.00 & 50.00 & 50.00
& 50.00 & 50.00 & 50.00
& 50.00 & 50.00 & 50.00 \\

\bottomrule
\end{tabular}
}

\vspace{-0.1in}
\caption{\textbf{Model performance on the scale-invariance task aggregated across all scales.} All models achieve near-perfect performance on Times New Roman and Handwritten English characters, indicating robustness to scale changes. In contrast, performance on Omniglot is substantially lower and exhibits greater variation in recall (TPR) and specificity (TNR) across models.}
\vspace{-0.8em}
\label{tab:scale_results}
\end{table}
\noindent\textbf{Findings:} 
From Table~\ref{tab:scale_results}, we observe that Times New Roman achieves near-perfect performance ($>98\%$) across all models, with both recall and specificity remaining high, indicating that models reliably identify characters across scales. Handwritten English follows closely, with all models exceeding $95\%$ accuracy. Performance drops substantially for Omniglot scripts, with accuracy ranging from $74.21\%$ ({\qwenThirtyTwo}) to $82.56\%$ ({\gemini}). Within Omniglot, for \qwenSeven, more familiar scripts such as Greek (TPR = $70.83\%$) achieve substantially higher recall than less familiar scripts such as Braille (TPR = $3.85\%$), while specificity remains near $100\%$ (more in Suppl.). Overall, these results indicates that the performance is strongly driven by script familiarity, with models reliably recognizing scaled variants of familiar scripts but performing poorly on less familiar ones. 
\subsection{Understanding VLMs fragility to simple transformations} \label{sec:why}

\begin{tcolorbox}[
  enhanced,
  colback=white,
  colframe=black,
  boxrule=0.6pt,
  arc=1.5mm,
  title=Thinking trace comparison for {\gemini},
  fonttitle=\bfseries\footnotesize,
  coltitle=white,
  colbacktitle=black,
  left=2pt,right=2pt,top=2pt,bottom=2pt
]
\label{fig:thinking}

{\scriptsize
\textbf{Rotation prompt:} If I rotate the first image, can I get the second image? Answer in curly brackets, e.g. \{Yes\} or \{No\}.
\par\vspace{2pt}
}

\noindent
\begin{minipage}[t]{0.49\linewidth}
\vspace{0pt}
\begin{tcolorbox}[
  enhanced,
  colback=white,
  colframe=black,
  boxrule=0.3pt,
  arc=1mm,
  left=2pt,right=2pt,top=2pt,bottom=2pt
]
\tiny

\noindent
\begin{minipage}[t]{0.72\linewidth}
\vspace{0pt}
\textcolor{scriptblue}{Times New Roman}\par\vspace{2pt}

{\itshape
\textbf{Thinking trace:} \\
\textbf{Assessing the Images}\par
I've determined both images represent forms of ``C'' and are now noting stylistic differences.\par\vspace{2pt}

\textbf{Confirming Image Similarity}\par
The images are confirmed as rotations of a `C', with careful serif and thickness matching, confirming the 90 degree rotation.
}\par\vspace{2pt}

\textbf{Response:} \{Yes\}
\end{minipage}
\hfill
\begin{minipage}[t]{0.25\linewidth}
\vspace{0pt}
\centering
\includegraphics[width=0.65\linewidth]{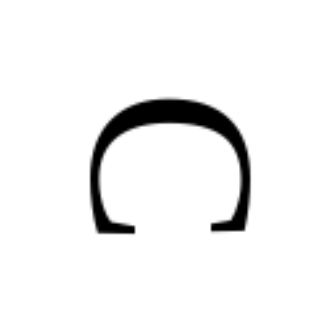}\\[-2pt]
\includegraphics[width=0.65\linewidth]{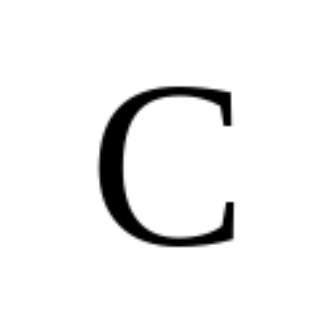}
\end{minipage}

\end{tcolorbox}
\end{minipage}\hfill
\begin{minipage}[t]{0.49\linewidth}
\vspace{0pt}
\begin{tcolorbox}[
  enhanced,
  colback=white,
  colframe=black,
  boxrule=0.3pt,
  arc=1mm,
  left=2pt,right=2pt,top=2pt,bottom=2pt
]
\tiny

\noindent
\begin{minipage}[t]{0.6\linewidth}
\vspace{0pt}
\textcolor{scriptorange}{Omniglot}\par\vspace{2pt}

{\itshape
\textbf{Thinking trace:} \\
\textbf{Assessing the Images}\par
I've initiated the analysis, aiming to ascertain the rotation required for image comparison and deliver a definitive answer.\par\vspace{2pt}

\textbf{Comparing Character Forms}\par
I've examined the images' structure, tracing each character, and tested rotations without finding a match, deeming them likely different.
}\par\vspace{2pt}

\textbf{Response:} \{No\}
\end{minipage}
\hfill
\begin{minipage}[t]{0.25\linewidth}
\vspace{0pt}
\centering
\includegraphics[width=0.65\linewidth]{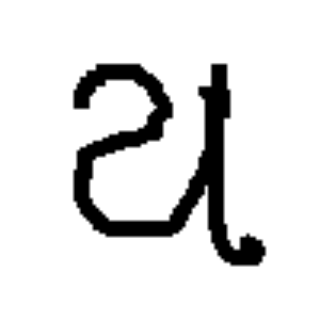}\\[-2pt]
\includegraphics[width=0.65\linewidth]{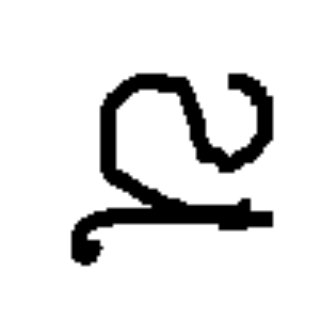}
\end{minipage}

\end{tcolorbox}
\end{minipage}

\end{tcolorbox}
To this end, we examine Gemini-2.5-Pro's~\citep{gemini} reasoning traces on $90^\circ$ rotation tasks (Sec.~\ref{sec:trans_inv}) by comparing familiar Times New Roman characters with unfamiliar Gujarati (Omniglot) characters. For the familiar character ``C,'' notice that the model identifies the letter's identity \textit{before} confirming the rotation. Conversely, for the unfamiliar Gujarati character, the model adopts a more low-level structural analysis of the character's geometry. We make a similar observation upon analyzing thinking traces of several other unfamiliar scripts. While this shift to structural analysis on unfamiliar scripts mirrors how humans \textit{might} solve a similar problem, Gemini-2.5-Pro's subsequent performance drop reveals a critical deficiency: current \textit{VLMs lack the robust geometric reasoning required to succeed without semantic anchors.} 

\begin{table}[!t]
\vspace{-1.2em}
\centering
\LARGE

\resizebox{0.99\textwidth}{!}{%
\begin{tabular}{llcccccccccccc}
\toprule
& 
& \multicolumn{2}{>{\columncolor{cyan!10}}c}{\textbf{\qwenSeven}} 
& \multicolumn{2}{>{\columncolor{cyan!10}}c}{\textbf{\qwenThirtyTwo}}
& \multicolumn{2}{>{\columncolor{blue!10}}c}{\textbf{\qwenThreeEight}}
& \multicolumn{2}{>{\columncolor{blue!10}}c}{\textbf{\qwenThreeThirty}}
& \multicolumn{2}{>{\columncolor{gray!10}}c}{\textbf{\gpt}}
& \multicolumn{2}{>{\columncolor{gray!10}}c}{\textbf{\gemini}} \\
\cmidrule(lr){3-4}
\cmidrule(lr){5-6}
\cmidrule(lr){7-8}
\cmidrule(lr){9-10}
\cmidrule(lr){11-12}
\cmidrule(lr){13-14}

\textbf{Script} & \textbf{ICL Setting}
& \textbf{TNR} & \textbf{\tprhead}
& \textbf{TNR} & \textbf{\tprhead}
& \textbf{TNR} & \textbf{\tprhead}
& \textbf{TNR} & \textbf{\tprhead}
& \textbf{TNR} & \textbf{\tprhead}
& \textbf{TNR} & \textbf{\tprhead} \\
\midrule

\multirow{3}{*}{Malayalam (top-tier)}
& None
& 100.00 & 00.00\same
& 97.87 & 06.38\same
& 100.00 & 00.00\same
& 100.00 & 02.13\same
& 93.62 & 34.04\same
& 100.00 & 38.30\same \\

& Few-shot
& 100.00 & 00.00\same
& 65.96 & 51.06\up
& 100.00 & 34.04\up
& 78.72 & 72.34\up
& 91.49 & 85.11\up
& 97.87 & \textcolor{red}{34.04}\down \\

& Rotational Grid
& 100.00 & 00.00\same
& 91.49 & 12.77\up
& 65.96 & \best{65.96}\up
& 23.40 & \best{87.23}\up
& 72.34 & \best{97.87}\up
& 93.62 & \best{48.94}\up \\

\noalign{\vskip 3pt}
\cdashline{1-14}
\noalign{\vskip 2pt}

\multirow{3}{*}{Tengwar (medium-tier)}
& None
& 100.00 & 00.00\same
& 88.00 & 16.00\same
& 100.00 & 00.00\same
& 100.00 & 00.00\same
& 100.00 & 20.00\same
& 96.00 & 32.00\same \\

& Few-shot
& 100.00 & 00.00\same
& 68.00 & \best{76.00}\up
& 100.00 & 16.00\up
& 68.00 & 72.00\up
& 80.00 & 68.00\up
& 92.00 & 32.00\same \\

& Rotational Grid
& 100.00 & 00.00\same
& 88.00 & 24.00\up
& 72.00 & \best{88.00}\up
& 40.00 & \best{76.00}\up
& 60.00 & \best{88.00}\up
& 88.00 & \best{56.00}\up \\

\noalign{\vskip 3pt}
\cdashline{1-14}
\noalign{\vskip 2pt}

\multirow{3}{*}{Braille (bottom-tier)}
& None
& 100.00 & 00.00\same
& 65.38 & 15.38\same
& 100.00 & 00.00\same
& 100.00 & 03.85\same
& 100.00 & 07.69\same
& 100.00 & \best{50.00}\same \\

& Few-shot
& 100.00 & 00.00\same
& 88.46 & \best{19.23}\up
& 100.00 & 11.54\up
& 92.31 & \best{50.00}\up
& 92.31 & \best{73.08}\up
& 96.15 & \textcolor{red}{26.92}\down \\

& Rotational Grid
& 100.00 & 00.00\same
& 100.00 & \textcolor{red}{00.00}\down
& 100.00 & \best{11.54}\up
& 73.08 & \best{50.00}\up
& 69.23 & 65.38\up
& 100.00 & \textcolor{red}{46.15}\down \\

\bottomrule
\end{tabular}}

\vspace{-0.4em}
\caption{\textbf{Model performance with in-context learning and structured visual prompting across scripts.} Arrows indicate change in TPR relative to the \textit{None} setting: \textcolor{teal}{\(\uparrow\)} denotes improvement and \textcolor{red}{\(\downarrow\)} denotes degradation. 
Both few-shot prompting and rotational grid inputs increase TPR across models, but often reduce TNRs. Improvements are larger for higher-capacity models. 
}
\vspace{-0.8em}
\label{tab:icl_final}
\end{table}
\subsection{Can transformation invariance be instilled?} \label{sec:icl}

\noindent\textbf{In-Context Learning (ICL).} Having established that MLLMs fail systematically on transformation tasks, we now investigate if in-context learning (ICL)~\citep{icl} and structured visual prompting can mitigate these errors. We evaluate across three representative Omniglot scripts, selected based on their performance on the rotation invariance task (Sec.~\ref{sec:rot}): Malayalam (top-tier), Tengwar (medium-tier), and Braille (bottom-tier). 
To the system prompt, we prepend two labeled examples: a positive pair with the caption ``\textit{This is Image B, which is a rotated version of Image A}," and a negative pair labeled ``\textit{This is Image D, which is NOT a rotated version of Image C.}" The positive example uses Angelic script characters (Images A and B); the negative uses an Angelic character as \textit{Image C} and a Gujarati character as the rotated variant (\textit{Image D}) (see suppl. material).

\noindent \textbf{Findings:}  From Table~\ref{tab:icl_final}, ICL improves performance primarily through increases in TPR, even with only \textbf{two} example pairs. For instance, {\qwenThirtyTwo} improves from $6.38\%$ to $51.06\%$ on Malayalam, while {\gpt} improves from $34.04\%$ to $85.11\%$. However, these gains often come at the cost of reduced TNR (e.g., {\qwenThirtyTwo} drops from $97.87\%$ to $65.96\%$ for Malayalam), indicating increased false positives. Improvements remain limited for smaller models such as {\qwenSeven}, where TPR remains zero.

\noindent\textbf{Rotational Grid.} As done in RotBench~\citep{Rotbench}, we construct a single composite image showing a character at four rotation angles: $0\degree$, $90\degree$, $180\degree$, $270\degree$. We use characters from Angelic and Gujarati, both mid-performing scripts, to construct the grid. 
We then prepend this image and a prompt describing the grid (see suppl. material) with the rotation task prompt in Sec.~\ref{sec:Omniglotprompt}.

\noindent \textbf{Findings:} From Table~\ref{tab:icl_final}, rotational grid increases TPR but causes larger drops in TNR than ICL. For example, the TPR of {\qwenThreeEight} improves from $0.00\%$ to $65.96\%$ on Malayalam, but TNR drops from $100.00\%$ to $65.96\%$. Similarly, {\qwenThreeThirty} improves from $2.13\%$ to $87.23\%$, at the cost of TNR plummeting to $23.40\%$. Even {\gpt} follows this pattern, achieving a near-perfect $97.87\%$ TPR but with a degraded TNR of $72.34\%$.

From these two experiments, it is clear that both approaches help instill \textit{some} but not a robust understanding of rotation recognition. While we hypothesized that ICL and the rotational grid would provide the visual evidence necessary to map the input across different angles, it appears to make the high-capacity models ``over-eager'' and induce a confirmation bias that spikes TPR at the expense of discriminative accuracy.

\section{Conclusion}
This work disentangles the geometric reasoning and semantic familiarity of state-of-the-art vision-language models. Across three transformations, four visual domains, and six models, we demonstrate that while performance is robust on semantically rich, familiar inputs (e.g., natural images), it degrades sharply on sketches, symbolic characters, and unfamiliar scripts. These failures persist regardless of model architecture, scale, or prompting strategy, and are only partially mitigated by in-context learning or structured visual prompts.
Our findings reveal that current VLMs lack a fundamental, invariant grasp of geometry and instead rely on ``semantic anchors'' to navigate spatial tasks. Future research must move beyond surface-level data familiarity, exploring architectural innovations and targeted data augmentations that instill true, zero-shot geometric reasoning capabilities. 

\noindent\textbf{Limitations.} While our benchmark isolates fundamental visual failures, it focuses exclusively on 2D transformations. We do not evaluate 3D rotations, which introduce complex viewpoint changes. Furthermore, because we employ a targeted diagnostic approach, we do not directly measure how these geometric deficits impact broader downstream applications, such as visual grounding or embodied robotics.

\noindent\textbf{Acknowledgments.} We thank Thomas Fel for especially helpful discussions. We also thank Dana Arad, David Bau, Mark Crovella, and Arsha Nagrani for helpful discussions and feedback, and, Dahye Kim, Manushree Vasu, and Chaitanya Chakka from our research group at BU for helpful discussions and feedback.

\newpage

\bibliographystyle{colm2026_conference}
\bibliography{main}

\newpage

\clearpage
\appendix

\section*{Supplementary Material for ``Semantic Richness or Geometric Reasoning? The Fragility of VLM’s Visual Invariance"}

\section*{Table of Contents}

\begin{tabular}{@{}p{0.08\linewidth}p{0.86\linewidth}@{}}

\textbf{A.} & \hyperref[sec:supp-datasets]{Dataset Examples} \dotfill \pageref{sec:supp-datasets} \\

& \hspace{1em} \textbf{A.1} \hyperref[sec:supp-datasets-over]{Dataset overview} \dotfill \pageref{sec:supp-datasets-over} \\
& \hspace{1em} \textbf{A.2} \hyperref[sec:supp-omniglot]{Omniglot Dataset} \dotfill \pageref{sec:supp-omniglot} \\
& \hspace{1em} \textbf{A.3} \hyperref[sec:supp-PACS]{PACS Dataset} \dotfill \pageref{sec:supp-PACS} \\
& \hspace{1em} \textbf{A.4} \hyperref[sec:supp-times]{Times New Roman Dataset} \dotfill \pageref{sec:supp-times} \\
& \hspace{1em} \textbf{A.5} \hyperref[sec:supp-hand]{Handwritten English Dataset} \dotfill \pageref{sec:supp-hand} \\[0.5ex]

\textbf{B.} & \hyperref[sec:supp-data]{Preprocessing and Transformations} \dotfill \pageref{sec:supp-data} \\

\textbf{C.} & \hyperref[sec:supp-enc]{Vision encoders studied for rotational invariance} \dotfill \pageref{sec:supp-enc} \\

\textbf{D.} & \hyperref[sec:supp-decoder]{Interaction with the Language Decoder} \dotfill \pageref{sec:supp-decoder} \\

\textbf{E.} & \hyperref[sec:supp-scale]{PACS Performance for the Scale Invariance Task} \dotfill \pageref{sec:supp-scale} \\[0.3ex]

\textbf{F.} & \hyperref[sec:supp-per]{Perimetric Complexity Analysis} \dotfill \pageref{sec:supp-per} \\

\textbf{G.} & \hyperref[sec:supp-scale-add]{Additional Scale Invariance Analysis} \dotfill \pageref{sec:supp-scale-add} \\

\textbf{H.} & \hyperref[sec:supp-icl-setup]{ICL and Rotational Grid Setup} \dotfill \pageref{sec:supp-icl-setup} \\

& \hspace{1em} \textbf{H.1} \hyperref[sec:supp-icl-fewshot]{Few-Shot Setup} \dotfill \pageref{sec:supp-icl-fewshot} \\
& \hspace{1em} \textbf{H.2} \hyperref[sec:supp-rot-grid]{Rotational Grid Setup} \dotfill \pageref{sec:supp-rot-grid} \\

\textbf{I.} & \hyperref[sec:supp-full]{Additional Results} \dotfill \pageref{sec:supp-full} \\

& \hspace{1em} \textbf{I.1} \hyperref[sec:supp-rot-char]{Rotation recognition performance across character datasets} \dotfill \pageref{sec:supp-rot-char} \\
& \hspace{1em} \textbf{I.2} \hyperref[sec:linear_probe]{Linear Probing of Rotation Information} \dotfill \pageref{sec:linear_probe} \\

& \hspace{1em} \textbf{I.3} \hyperref[sec:supp-rot-pacs-0-90]{PACS rotation results for $10^\circ$--$90^\circ$} \dotfill \pageref{sec:supp-rot-pacs-0-90} \\
& \hspace{1em} \textbf{I.4} \hyperref[sec:supp-identity-pacs]{PACS identity experiment results} \dotfill \pageref{sec:supp-identity-pacs} \\

& \hspace{1em} \textbf{I.5} \hyperref[sec:supp-icl]{Model performance with In-Context Learning} \dotfill \pageref{sec:supp-icl} \\

& \hspace{1em} \textbf{I.6} \hyperref[sec:prompt_analysis]{Investigating Response Bias and Prompt Sensitivity} \dotfill \pageref{sec:prompt_analysis} \\
& \hspace{1em} \textbf{I.7} \hyperref[sec:human_study]{Human Baseline Study} \dotfill \pageref{sec:human_study} \\
& \hspace{1em} \textbf{I.8} \hyperref[sec:stats_robustness]{Statistical Robustness and Confidence Intervals} \dotfill \pageref{sec:stats_robustness} \\
& \hspace{1em} \textbf{I.9} \hyperref[sec:additional_models]{Evaluating Additional Open-Source Models} \dotfill \pageref{sec:additional_models} \\
& \hspace{1em} \textbf{I.10} \hyperref[sec:data_domain_grounding]{The Role of Data Domain and Semantic Grounding} \dotfill \pageref{sec:data_domain_grounding} \\
\end{tabular}

\newpage
\clearpage

\section{Dataset Examples}
\label{sec:supp-datasets}

\subsection{Dataset overview}\label{sec:supp-datasets-over}

\begin{figure}[!h]
  \centering
  \includegraphics[width=0.75\linewidth, keepaspectratio]{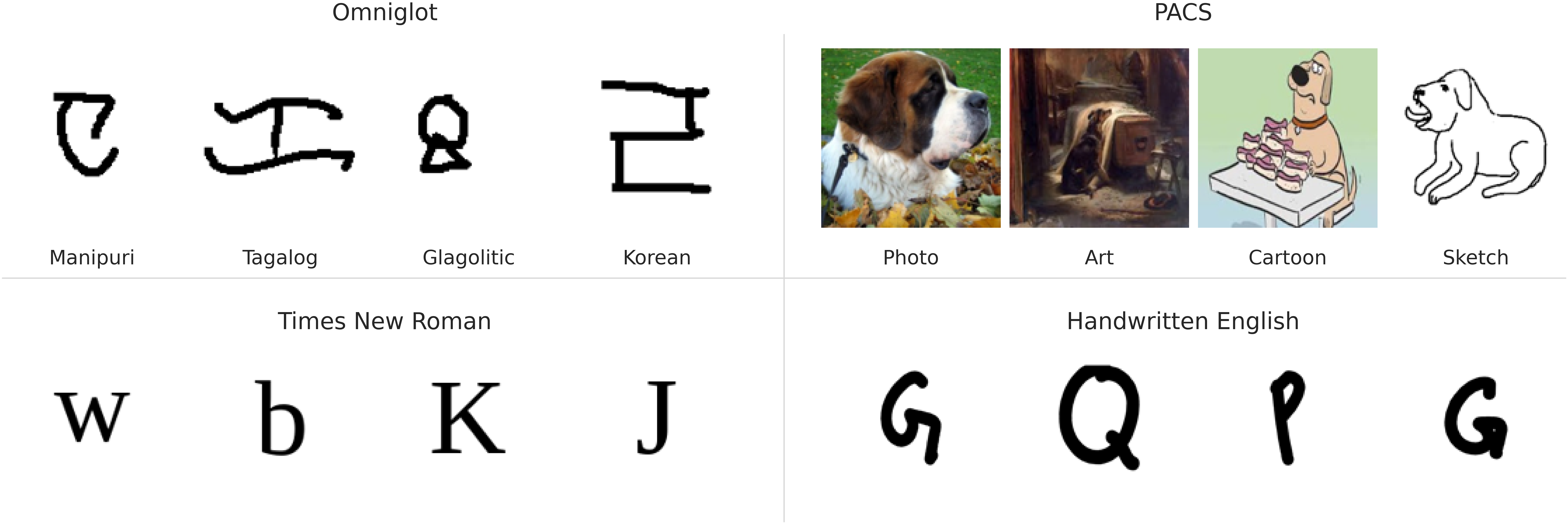}
  \vspace{-0.8em}
  \caption{\textbf{Datasets used in our evaluation.}
    Omniglot~\citep{omniglot} contains handwritten binary characters from $50$ diverse scripts. Times New Roman~\citep{timesnewroman} provides standardized English characters rendered in a fixed typeface. Handwritten English~\cite{handwritten_english_characters_digits} includes handwritten characters from the English alphabet. PACS~\citep{PACS} contains images of common object categories (e.g., guitar, dog, elephant) across four visual domains: \textbf{P}hotograph, \textbf{A}rt, \textbf{C}artoon, and \textbf{S}ketch. 
    Together, these datasets allow us to evaluate transformation invariance in MLLMs across scripts, visual styles, and images with varying levels of semantic richness.}
  \label{fig:data_grid}
\end{figure}

\begin{table}[!h]
\centering
\scriptsize
\setlength{\tabcolsep}{6pt}
\begin{tabular}{lcc}
\toprule
\textbf{Dataset} & \textbf{\# Samples} & \textbf{Description} \\
\midrule
Omniglot & 1,623 & Handwritten characters across 50 scripts \\
Times New Roman & 52 & Printed English alphabet (A–Z, a–z) \\
Handwritten English & 52 & Handwritten English alphabet \\
PACS & 800 & 4 domains $\times$ 200 images each \\
\bottomrule
\end{tabular}
\caption{\textbf{Dataset overview.} Number of samples used for evaluation in each dataset.}
\label{tab:dataset_overview}
\end{table}

\begin{table}[!h]
\centering
\scriptsize
\setlength{\tabcolsep}{6pt}
\begin{tabular}{llll}
\toprule
\multicolumn{4}{c}{\textbf{Omniglot scripts}} \\
\midrule
Alphabet of the Magi & Armenian & Gujarati & Mongolian \\
Angelic & Asomtavruli (Georgian) & Gurmukhi & N'Ko \\
Anglo-Saxon Futhorc & Atemayar Qelisayer & Hebrew & Ojibwe (Canadian Aboriginal) \\
Arcadian & Atlantean & Inuktitut (Canadian Aboriginal) & Old Church Slavonic (Cyrillic) \\
Aurek-Besh & Avesta & Japanese (Hiragana) & Oriya \\
Balinese & Bengali & Japanese (Katakana) & Sanskrit \\
Blackfoot (Canadian Aboriginal) & Braille & Kannada & Sylheti \\
Burmese (Myanmar) & Cyrillic & Keble & Syriac (Estrangelo) \\
Early Aramaic & Futurama & Korean & Syriac (Serto) \\
Ge'ez & Glagolitic & Latin & Tagalog \\
Grantha & Greek & Malay (Jawi - Arabic) & Tengwar \\
Malayalam & Manipuri & Mkhedruli (Georgian) & Tibetan \\
Tifinagh & ULOG & & \\
\bottomrule
\end{tabular}
\caption{\textbf{Omniglot scripts.} List of 50 writing systems used in the Omniglot dataset.}
\label{tab:omniglot_scripts}
\end{table}

\begin{table}[!h]
\centering
\scriptsize
\begin{tabular}{ll}
\toprule
\textbf{Categories} & Dog, Elephant, Giraffe, Horse, Person, Guitar, House \\
\textbf{Domains} & Photograph, Art Painting, Cartoon, Sketch \\
\bottomrule
\end{tabular}
\caption{\textbf{PACS dataset.} Object categories and visual domains used in evaluation.}
\label{tab:pacs_summary}
\end{table}

\newpage
\clearpage

\subsection{Omniglot Dataset}
\label{sec:supp-omniglot}

\vspace*{\fill}
\begin{figure}[h]
  \centering
  \includegraphics[width=0.99\linewidth, keepaspectratio]{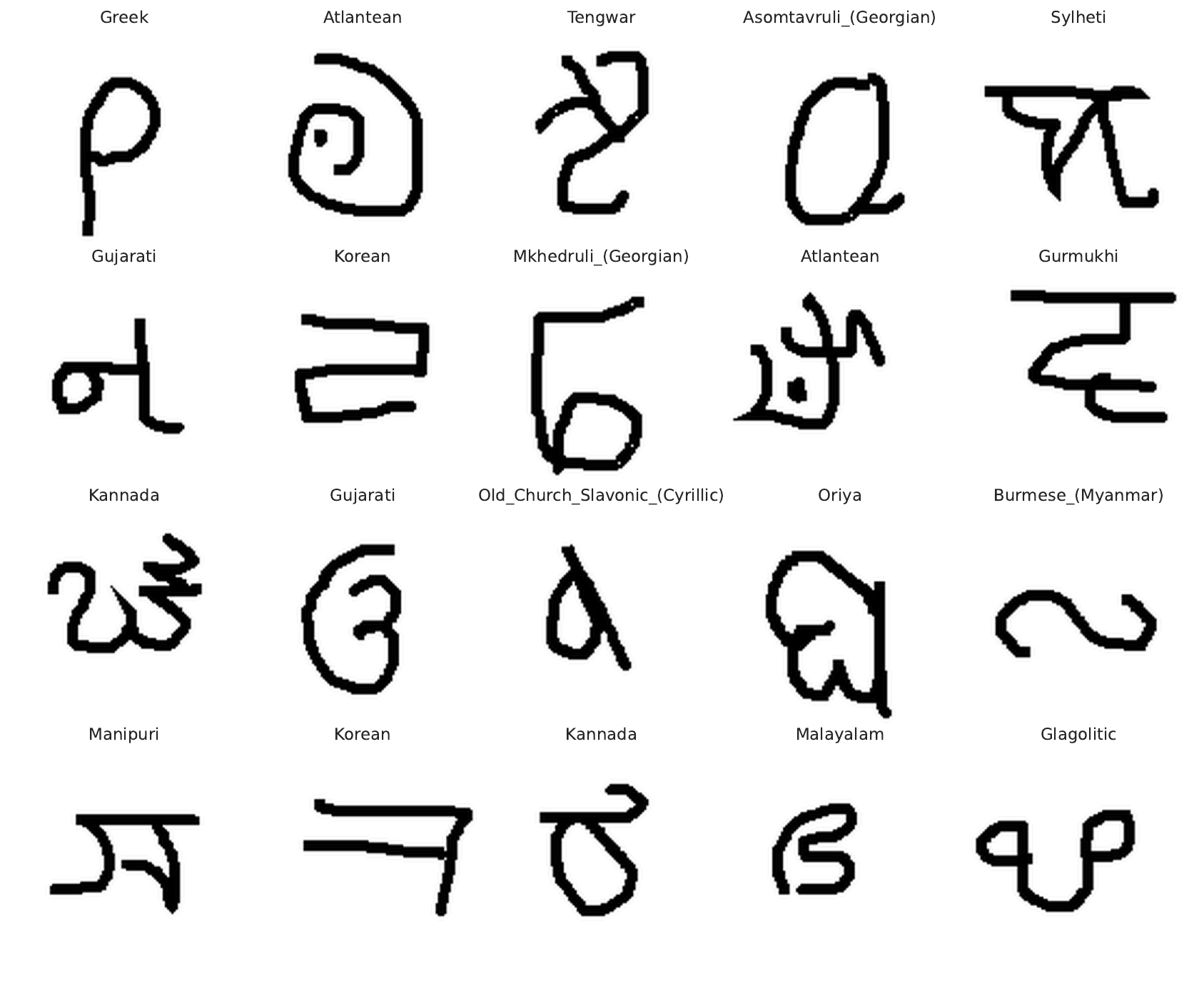}
  \caption{\textbf{Examples from the Omniglot dataset}}
  \label{fig:grid_omniglot}
\end{figure}
\vspace*{\fill}

\clearpage

\subsection{PACS Dataset}
\label{sec:supp-PACS}

\vspace*{\fill}
\begin{figure}[h]
  \centering
  \includegraphics[width=0.99\linewidth, keepaspectratio]{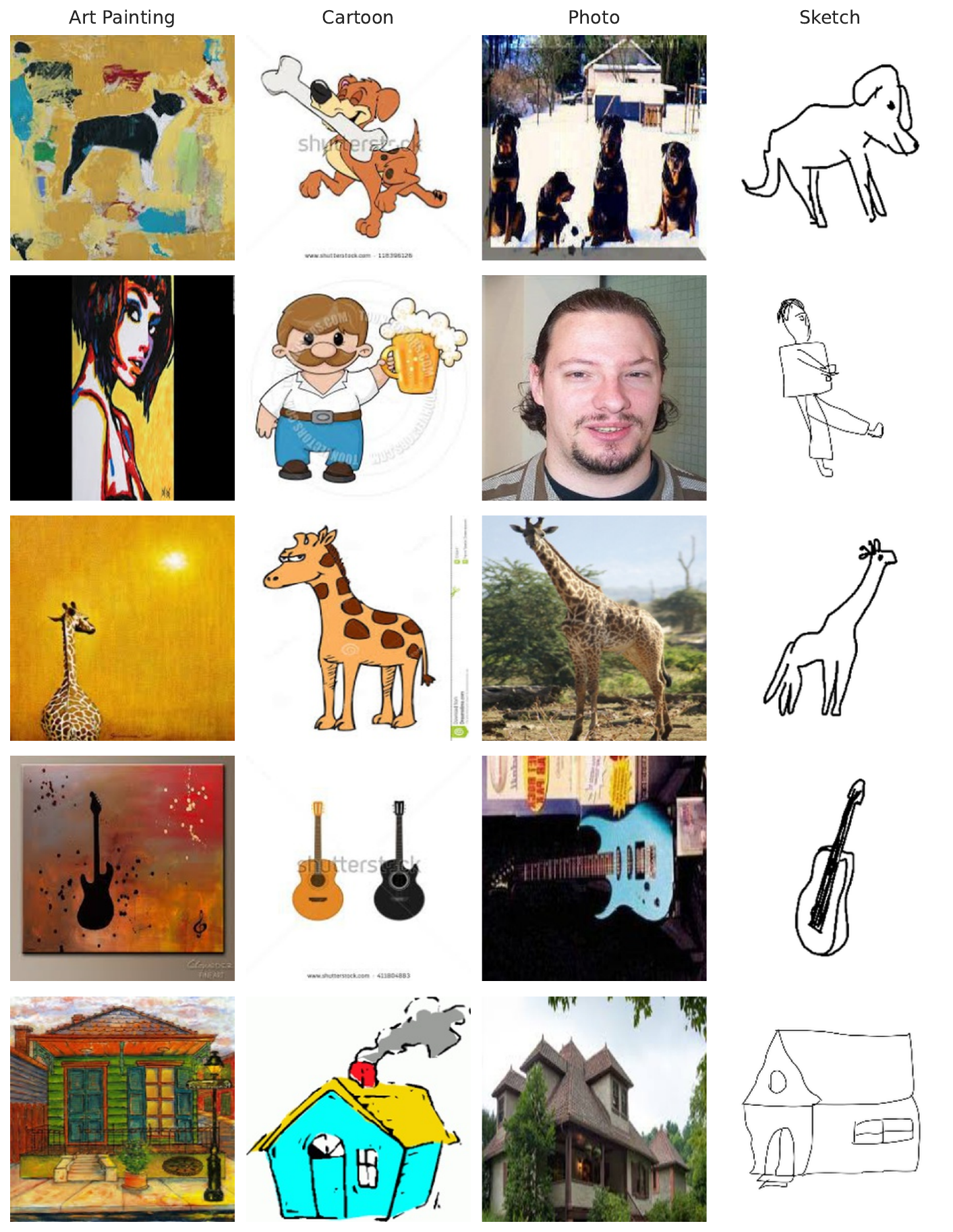}
  \caption{\textbf{Examples from the PACS dataset}}
  \label{fig:grid_pacs}
\end{figure}
\vspace*{\fill}

\clearpage

\subsection{Times New Roman Dataset}
\label{sec:supp-times}

\vspace*{\fill}
\begin{figure}[h]
  \centering
  \includegraphics[width=0.99\linewidth, keepaspectratio]{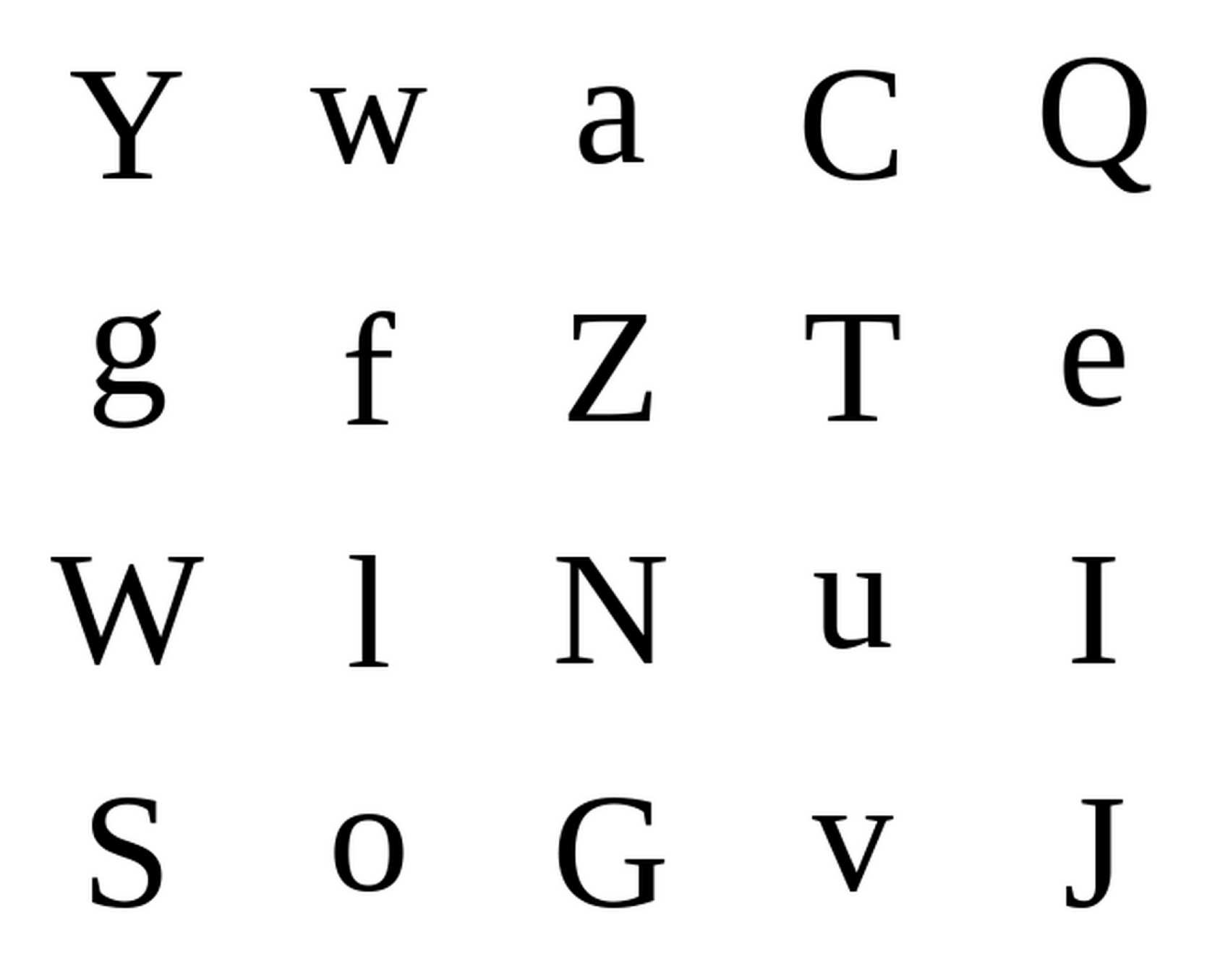}
  \caption{\textbf{Examples from the Times New Roman dataset}}
  \label{fig:grid_times}
\end{figure}
\vspace*{\fill}

\clearpage

\subsection{Handwritten English Dataset}
\label{sec:supp-hand}

\vspace*{\fill}
\begin{figure}[h]
  \centering
  \includegraphics[width=0.99\linewidth, keepaspectratio]{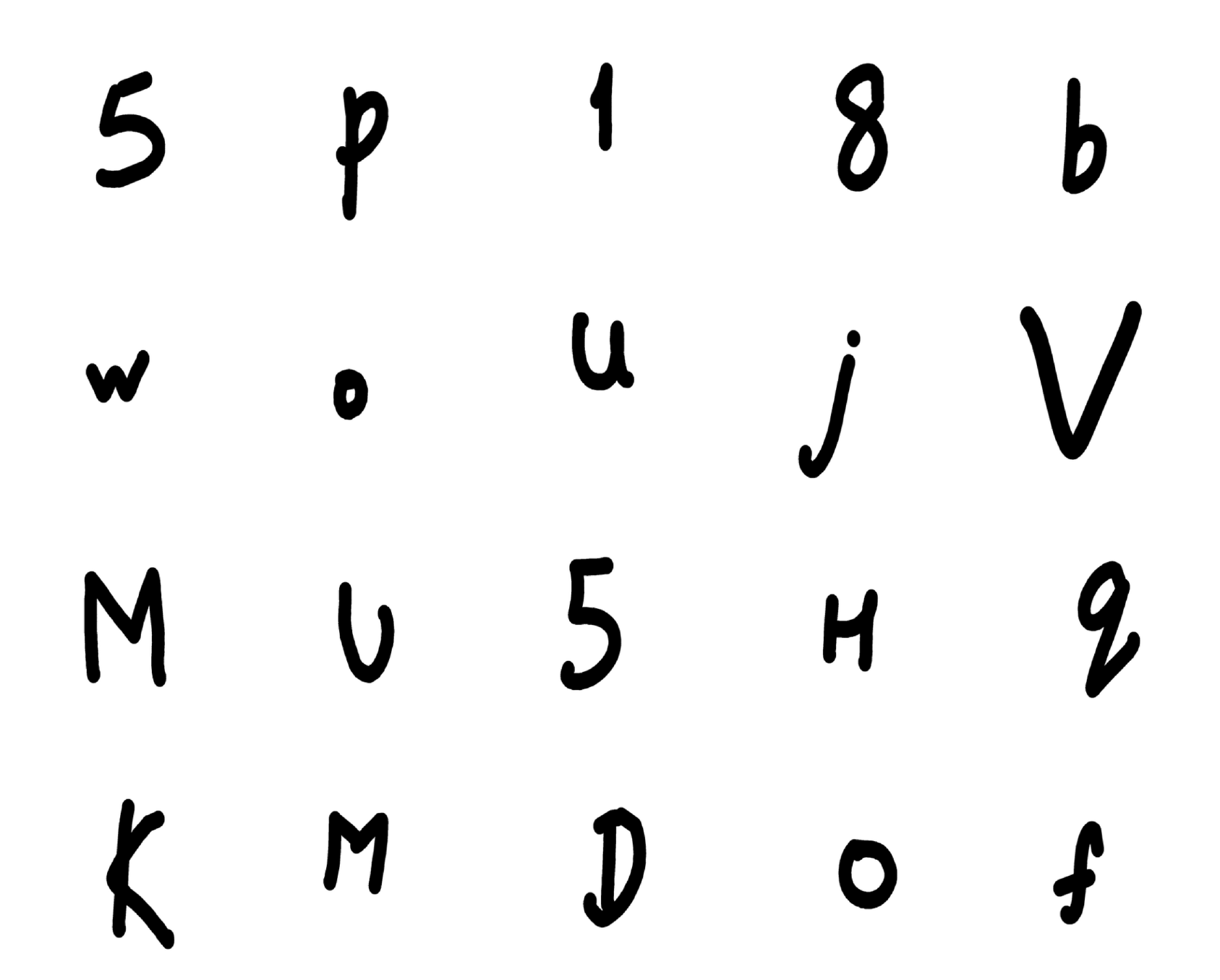}
  \caption{\textbf{Examples from the Handwritten English dataset}}
  \label{fig:grid_hand}
\end{figure}
\vspace*{\fill}

\clearpage

\section{Preprocessing and Transformations}
\label{sec:supp-data}

\begin{figure}[h]
  \centering
  \includegraphics[width=0.99\linewidth, keepaspectratio]{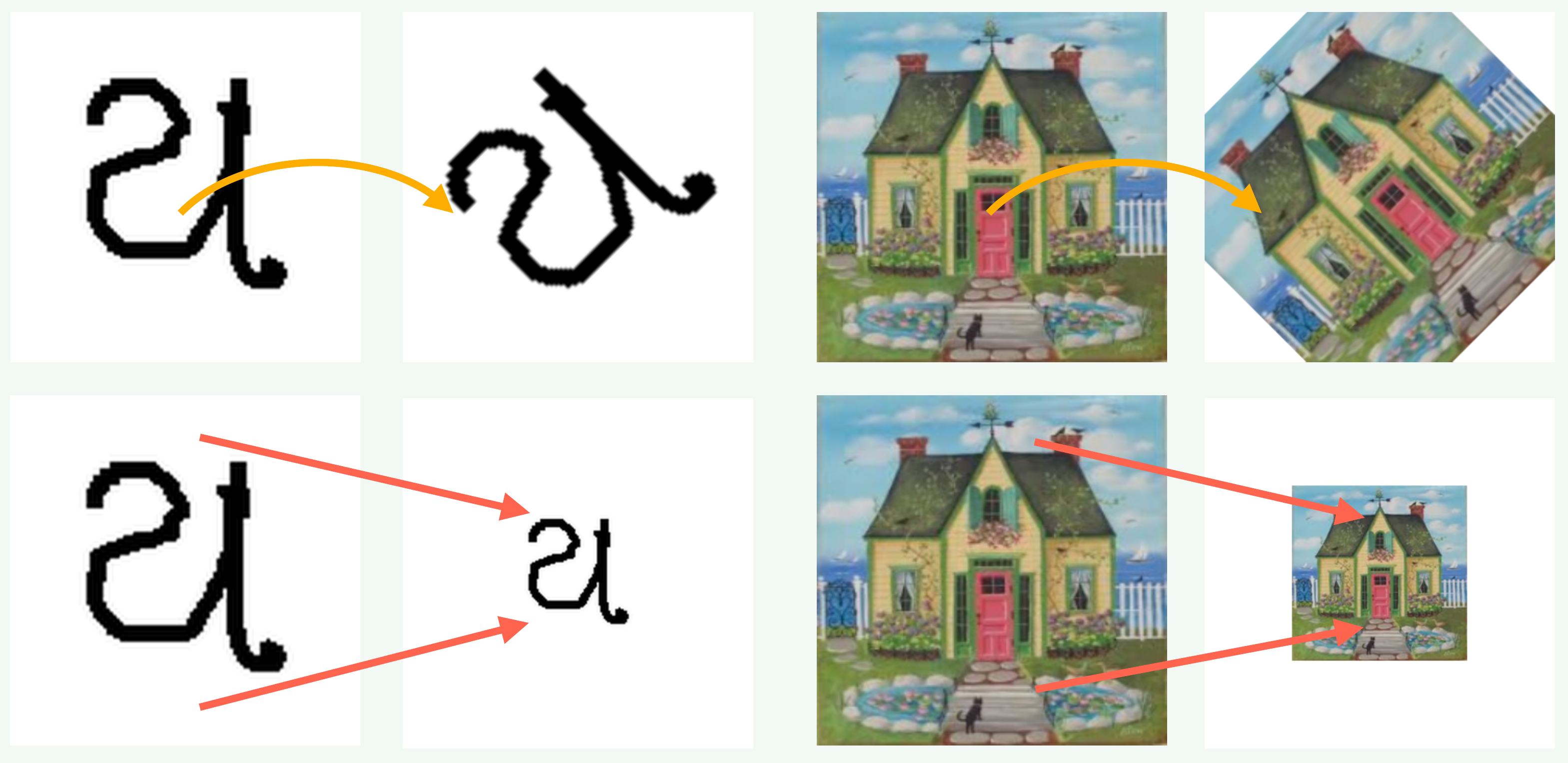}
  \caption{\textbf{Padding artifacts under rotation and scaling.} Top row shows rotation and bottom row shows scaling. Non-$90^\circ$ rotations introduce padding regions corresponding to pixels not covered by the original image. Scaling similarly introduces padding for resized images. For character datasets (left), padding matches the white background and is not visually salient. In contrast, for PACS (right), padding introduces visible artifacts under both rotation and scaling. Top row: Images rotated by $45^\circ$. Bottom row: Images scaled by $0.5$.}
  \label{fig:example}
\end{figure}

For rotation (Sec.~\ref{sec:rot}) and scale experiments (Sec.~\ref{sec:scale_exp}), transformations are applied about the image center. Non-orthogonal rotations introduce empty regions where no original pixels are present (Fig.~\ref{fig:example}); these regions are filled with white padding pixels for the character datasets (Omniglot, Times New Roman, and Handwritten English) and for PACS. 
For scale transformations, characters are resized by a factor $s \in \{0.1, 0.3, 0.5, 0.9\}$ and padded to maintain the original image dimensions (Fig.~\ref{fig:example}).
In the character datasets, images are already represented on a white background, so padding introduced by transformations is consistent with the background of the original images. In contrast, for PACS, scale transformations and non-orthogonal rotations introduce visible white padding regions. We acknowledge this as a dataset-dependent artifact that may influence model behavior. However, for the main analysis in Table~\ref{tab:pacs_results},  we restrict rotations to $90\degree$, $180\degree$, and $270\degree$, which avoids introducing such padding artifacts. 

\clearpage
\newpage

\section{Vision encoders studied for rotational invariance (Sec.~\ref{sec:enc_rot})}\label{sec:supp-enc}


\begin{table}[!h]
\centering
\resizebox{0.75\textwidth}{!}{%
\begin{tabular}{lcc}
\toprule
\textbf{Model} & \textbf{Feature Extraction} & \textbf{Dimension} \\
\midrule
CLIP (ViT-L/14)~\citep{clip} & CLS token & 768 \\
DINOv2 (ViT-L/14)~\citep{dino} & CLS token & 1024 \\
SigLIP (ViT-SO400M)~\citep{siglip} & MAP (multi-head attention pooling) & 1152 \\
Qwen2.5-VL-7B~\citep{qwen2} & Mean over vision tokens & 3584 \\
Stable Diffusion v2.1~\citep{stablediffusion} & Mean over layer 14 features & 1280 \\
\bottomrule
\end{tabular}}
\caption{\textbf{Vision feature extraction details.} For CLIP and DINOv2, we use the CLS token. For SigLIP, we use the multi-head attention pooling (MAP) output. For Qwen2.5-VL, we average vision token embeddings to obtain a global representation. For Stable Diffusion v2.1 we obtain features from layer $14$ in the UNet.}
\label{tab:feature_extraction}
\end{table}

\begin{figure}[h]
  \centering
  \includegraphics[width=0.99\linewidth, keepaspectratio]{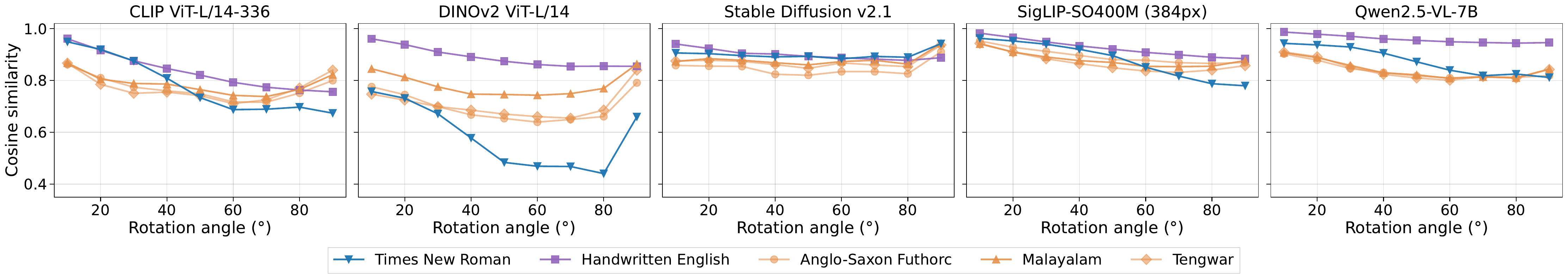}
  \caption{\textbf{Cosine similarity between features extracted from different vision encoders} on pairs of images under rotation. Select \textcolor{scriptorange}{Omniglot scripts} are shown in orange, while \textcolor{scriptblue}{Times New Roman} and \textcolor{scriptpurple}{Handwritten English} are shown in blue and purple respectively. Across all encoders, similarity decreases with increasing rotation angle, with DINOv2 showing the steepest drop and SigLIP and {\qwenSeven} maintaining relatively higher similarity.}
  \label{fig:grid_hand_app}
\end{figure}

\noindent\textbf{CLIP (ViT-L/14)~\citep{clip}}
We extract image features using the \texttt{[CLS]} token, which serves as a global representation of the input image. These features are used directly for cosine similarity computations.

\noindent\textbf{DINOv2 (ViT-L/14)~\citep{dino}}
We extract image features using the \texttt{[CLS]} token, which serves as a global representation of the input image. These features are used directly for cosine similarity computations.

\noindent\textbf{SigLIP (ViT-SO400M)~\citep{siglip}}
We use the multi-head attention pooling (MAP) output, which aggregates information across all visual tokens to form a global representation of the input image. 

\noindent\textbf{Qwen2.5-VL-7B~\citep{qwen2}}
We compute the mean of the vision token embeddings from the final layer to obtain a global image representation, which is then used for cosine similarity computations.

\noindent\textbf{Stable Diffusion v2.1~\citep{stablediffusion}}
Following prior work~\citep{kim2025revelio} showing that intermediate layers of diffusion models capture rich semantic information, we extract features from layer 14 of the U-Net encoder and compute their spatial mean to obtain a representative representation.

\clearpage
\newpage

\section{Suspect 4: Interaction with the Language Decoder}\label{sec:supp-decoder}

\begin{table}[!h]
\centering
\LARGE
\resizebox{0.90\columnwidth}{!}{
\begin{tabular}{lcccccccc}
\toprule
\textbf{Metric} 
& \multicolumn{1}{c}{\textcolor{scriptblue}{Times New Roman}}
& \multicolumn{1}{c}{\textcolor{scriptpurple}{Handwritten English}} 
& \multicolumn{1}{c}{\textcolor{scriptorange}{Braille}} 
& \multicolumn{1}{c}{\textcolor{scriptorange}{Balinese}} 
& \multicolumn{1}{c}{\textcolor{scriptorange}{Anglo-Saxon\_Futhorc}} 
& \multicolumn{1}{c}{\textcolor{scriptorange}{Tengwar}} 
& \multicolumn{1}{c}{\textcolor{scriptorange}{Malayalam}} 
& \multicolumn{1}{c}{\textcolor{scriptorange}{Grantha}} \\
\midrule

Acc. 
& 57.87 & 54.54 & 66.43 & 50.18 & 53.76 & 49.63 & 50.87 & 50.42 \\

TPR 
& 77.27 & 95.28 & 46.85 & 99.62 & 97.17 & 99.27 & 96.71 & 99.78 \\

TNR 
& 38.46 & 13.81 & 86.01 & 0.75 & 10.34 & 0.00 & 5.03 & 1.05 \\
\bottomrule
\end{tabular}
}
\vspace{-0.4em}
\caption{\textbf{Idefics2 accuracy performance on the rotation task across scripts.} 
\textcolor{scriptblue}{Times New Roman} and \textcolor{scriptpurple}{Handwritten English} are shown in blue and purple respectively, while \textcolor{scriptorange}{Omniglot scripts} are shown in orange. 
Accuracy remains near chance for most scripts, despite consistently high TPR and substantially lower TNR, indicating a strong ``YES'' prediction bias.}
\label{tab:llava_rotation_full}
\end{table}

The analyses in Sec.~\ref{sec:enc_rot} reveal a striking disconnect: vision encoders maintain high cosine similarity between a character and its rotated counterpart, yet MLLMs consistently fail to recognize the same rotation (Table~\ref{tab:llava_rotation_full}). 

To isolate the contribution of the language decoder, we evaluate Idefics2~\citep{idefics}, which uses a frozen {\siglip}~\citep{siglip} vision encoder, the same encoder evaluated in Sec.~\ref{sec:rot}. This enables a controlled comparison; if SigLIP preserves rotational similarity at the feature level but Idefics2 fails at the task level, the gap likely arises from how these representations are used within the MLLM (i.e., how the language decoder processes and utilizes those visual representations). Table~\ref{tab:llava_rotation_full} reports Idefics2's performance on the rotation task across scripts following the same setup as in Sec.~\ref{sec:rot}. Despite SigLIP maintaining high cosine similarity between rotated character pairs across all scripts (Fig.~\ref{fig:encoder_graphs_omniglot}), Idefics2 achieves only near-chance accuracy across all evaluated scripts ($\sim50$ – $66\%$). Performance varies widely across scripts and is driven by a strong “YES” bias: TPR is consistently high while TNR is often near zero, indicating frequent false positives. This mismatch shows that while the vision encoder produces highly similar representations for transformed inputs, this similarity alone is insufficient for reliable downstream reasoning about transformations.

\clearpage
\newpage

\section{PACS Performance for the Scale Invariance Task}
\label{sec:supp-scale}

\vspace*{\fill}
\begin{figure}[h]
  \centering
  \includegraphics[width=0.99\linewidth, keepaspectratio]{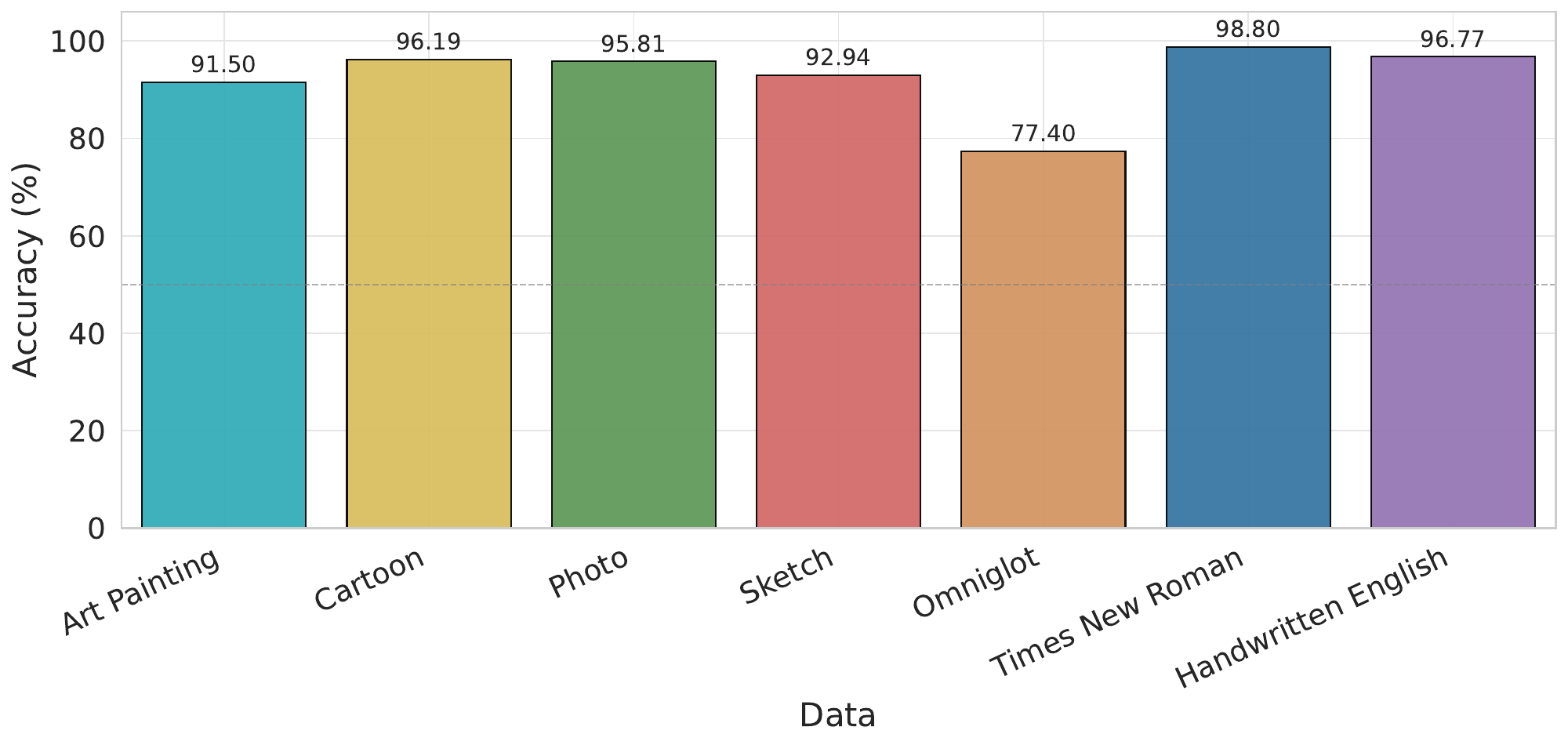}
  \caption{\textbf{Scale invariance performance across datasets for {\qwenTwo}.} aggregated over all scales (Sec.~\ref{sec:scale_exp}). Performance is high on natural image domains (Art Painting, Cartoon, Photo, Sketch) and familiar scripts (Times New Roman, Handwritten English), but drops on Omniglot, indicating reduced robustness under lower semantic familiarity.}
  \label{fig:grid_hand}
\end{figure}
\begin{figure}[h]
  \centering
  \includegraphics[width=0.99\linewidth, keepaspectratio]{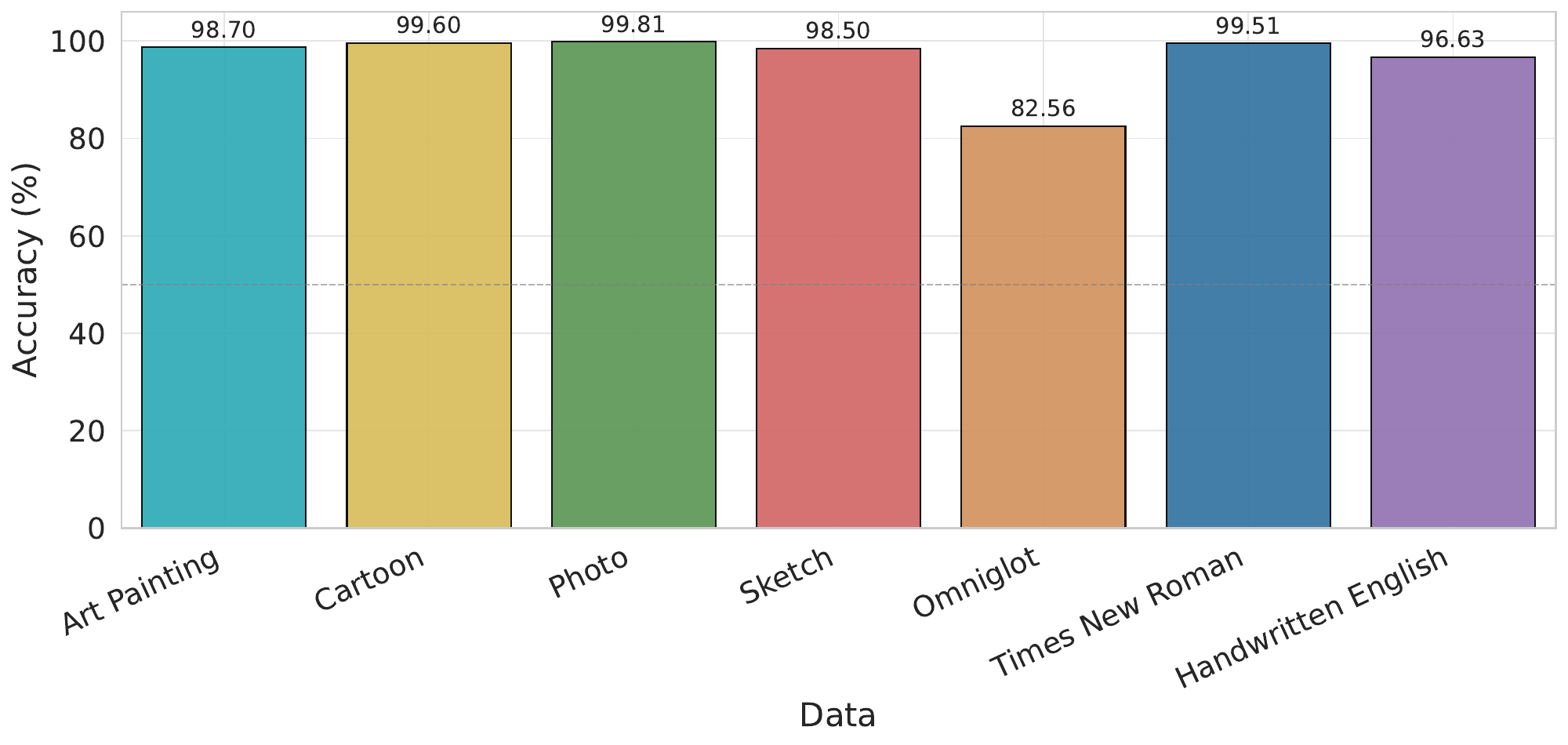}
  \caption{\textbf{Scale invariance performance across datasets for {\gemini}.} aggregated over all scales (Sec.~\ref{sec:scale_exp}). Performance is near-perfect on natural image domains (Art Painting, Cartoon, Photo, Sketch) and familiar scripts (Times New Roman, Handwritten English), but drops on Omniglot, indicating reduced robustness under lower semantic familiarity.}
  \label{fig:grid_hand}
\end{figure}
\vspace*{\fill}

\clearpage
\newpage

\section{Perimetric Complexity Analysis}
\label{sec:supp-per}


\begin{figure}[!h]
  \centering
  \includegraphics[width=0.8\linewidth]{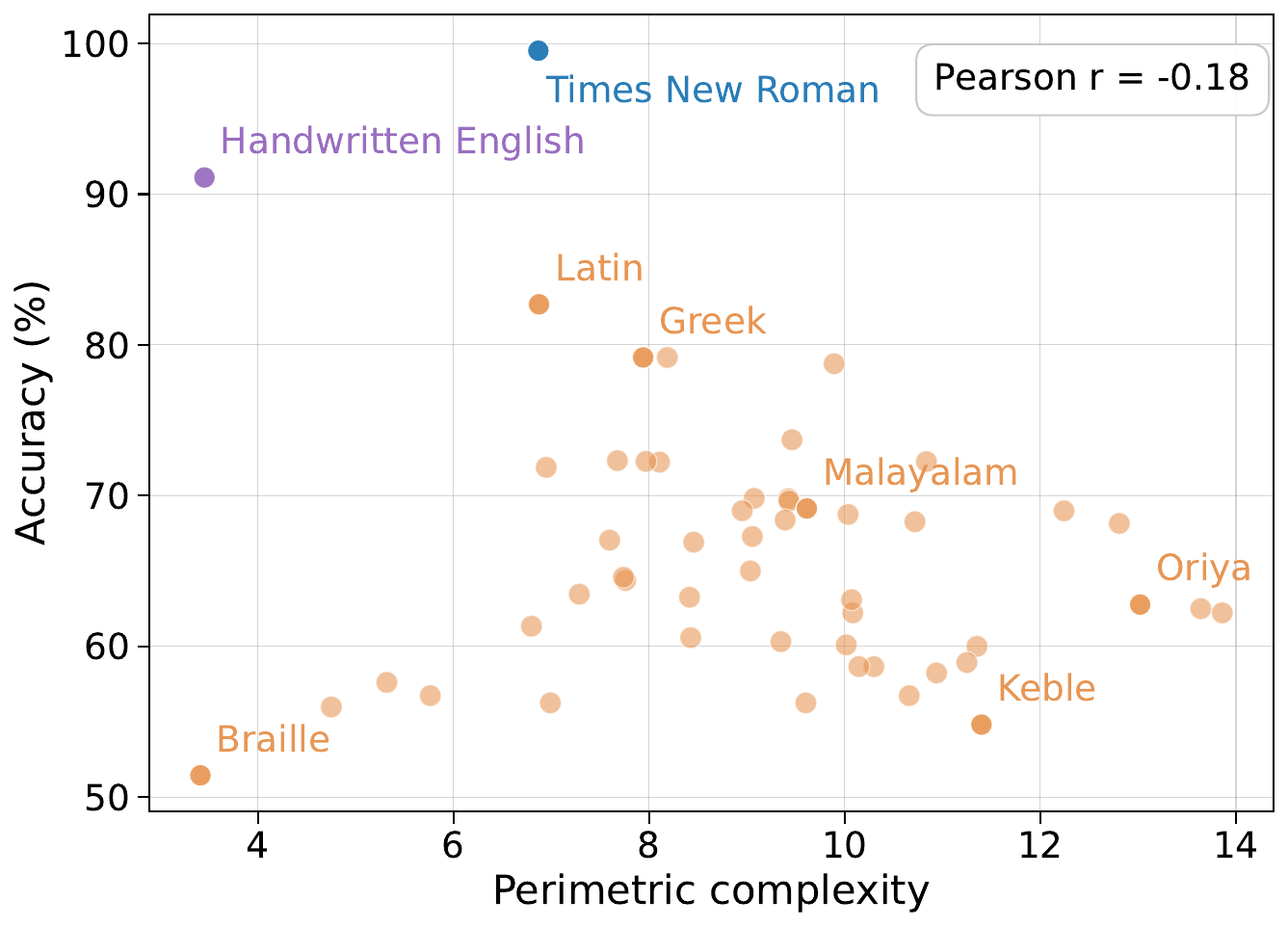}
  \caption{\textbf{Perimetric complexity vs performance.} Perimetric complexity shows a weak correlation ($r = -0.18$) with accuracy of {\qwenSeven} on the scale invariance task. \textcolor{scriptblue}{Times New Roman} and \textcolor{scriptpurple}{Handwritten English} are shown in blue and purple respectively, while \textcolor{scriptorange}{Omniglot scripts} are highlighted in orange.}
  \label{fig:perimetric_complexity}
\end{figure}

\noindent\textbf{Visual complexity alone does not account for performance differences across scripts.} We study if the performance struggle is more severe on more visually intricate and complex scripts. 
To test this hypothesis, we use perimetric complexity as used by~\cite{watson2011perimetric}, defined as $P^2 / A$, where $P$ is the total perimeter of the character and $A$ is the area occupied by the character. This scale- and orientation-invariant metric indicates structural intricacy. If structural complexity is the primary cause, scripts of higher
perimetric complexity would perform worse. However, the Pearson correlation between script-level complexity and mean accuracy is weak ($r = -0.18$), indicating that structural intricacy alone cannot explain the poor performance on unknown scripts. 

\clearpage
\newpage

\clearpage
\newpage

\section{Additional Scale Invariance Analysis}{\label{sec:supp-scale-add}}

\begin{figure}[!h]
  \centering
  \begin{subfigure}[t]{0.49\linewidth}
    \centering
    \includegraphics[width=\linewidth, keepaspectratio]
    {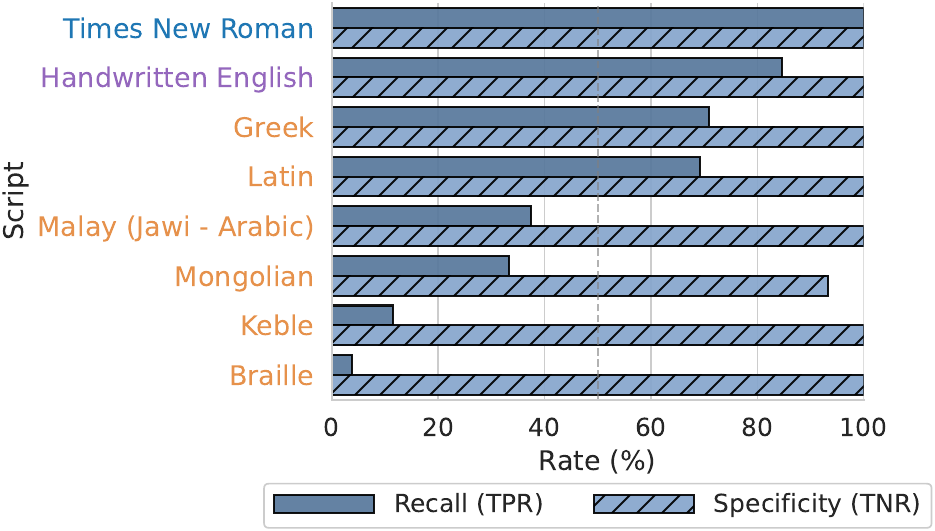}
    \caption{ \qwenSeven}
    \label{fig:scale_results_a}
  \end{subfigure}
  \hfill
  \begin{subfigure}[t]{0.49\linewidth}
    \centering
    \includegraphics[width=\linewidth, keepaspectratio]
    {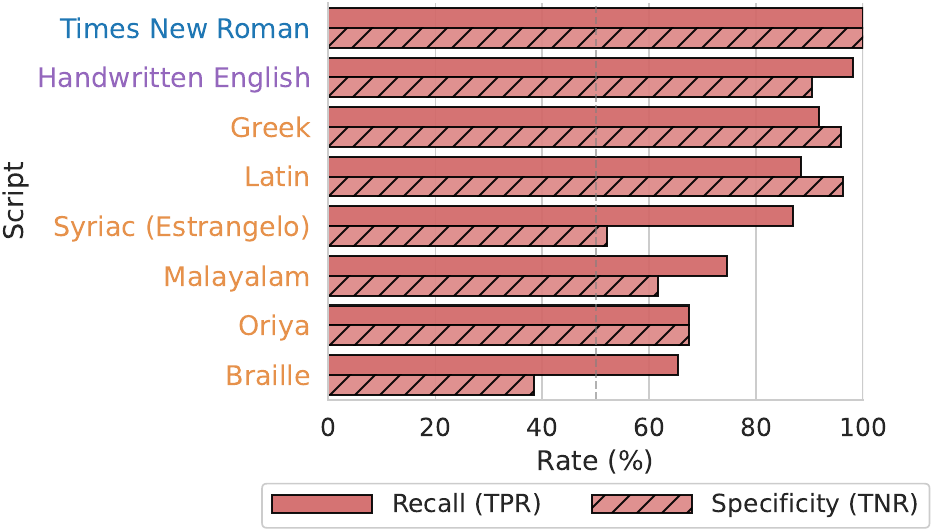}
    \caption{ \qwenThirtyTwo}
    \label{fig:scale_results_b}
  \end{subfigure}
\vspace{-0.8em}
  \caption{
\textbf{Recall and Specificity at scale $0.3\times$ for representative scripts for the scale-invariance task.}
\textcolor{scriptblue}{English characters rendered in Times New Roman}, 
\textcolor{scriptpurple}{Handwritten English characters}, and 
\textcolor{scriptorange}{Omniglot scripts} are shown in blue, purple, and orange respectively, and are selected to represent high-, medium-, and low-performing groups. Across both models, familiar scripts such as Greek 
and Latin consistently outperform less familiar scripts like Braille. While \qwenThirtyTwo~achieves higher recall than \qwenSeven~on low-performing Omniglot scripts, it exhibits lower specificity.}
  \label{fig:scale_results}
\end{figure}
\begin{figure}[!h]
  \centering
  \includegraphics[width=0.95\linewidth, keepaspectratio]{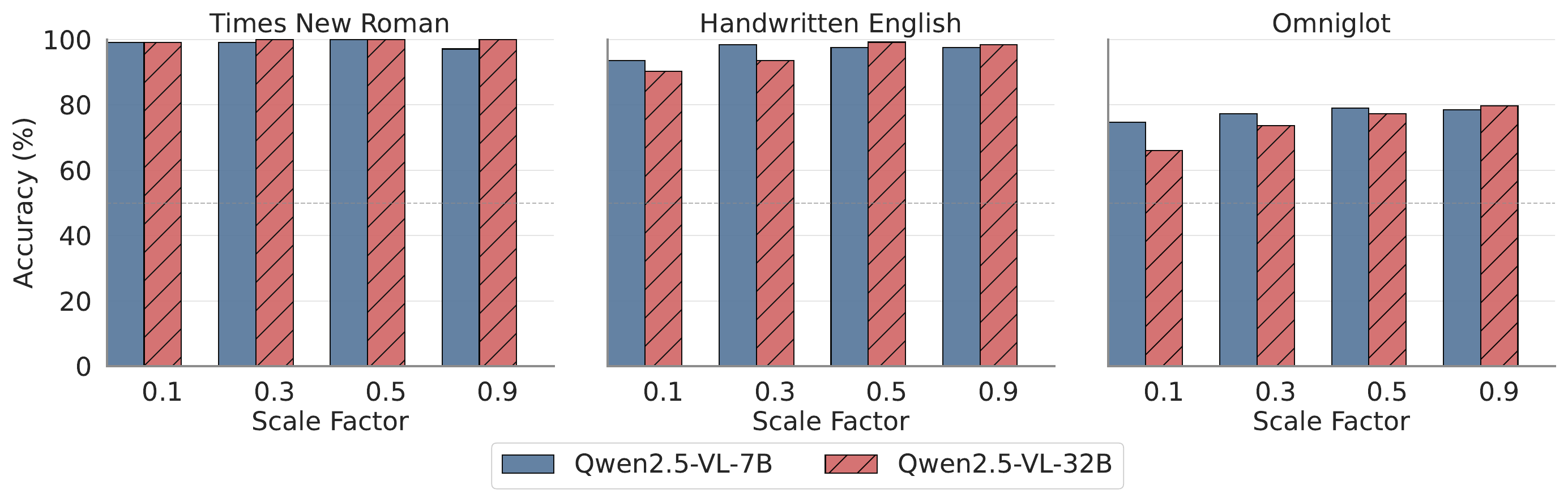}
  \caption{
\textbf{Model accuracy on the scale-invariance task across scale factors.
} Both {\qwenSeven} and {\qwenThirtyTwo} models maintain near-perfect accuracy for both Times New Roman and Handwritten English characters across all scales, while performance on Omniglot scripts is substantially and consistently lower for both models.}
  \label{fig:scale_model_comp}
\end{figure}

\clearpage
\newpage

\section{ICL and Rotational Grid Setup}
\label{sec:supp-icl-setup}
\subsection{Few-Shot Setup}\label{sec:supp-icl-fewshot}

\begin{figure}[h]
    \centering
    
    \fbox{
    \begin{minipage}{0.9\linewidth}
    \small
    \textbf{System Prompt:} \\
    You are a vision reasoning model. You will be shown pairs of images. 
    Determine whether one image is a rotated version of the other. \\
    
    \textbf{Examples:} \\
    ``This is Image B, which is a rotated version of Image A.'' \\
    ``This is Image D, which is NOT a rotated version of Image C.''
    \end{minipage}
    }
    
    \vspace{0.8em}
    
    \textbf{Positive Example} \\
    \begin{tabular}{cc}
        \includegraphics[width=0.22\textwidth]{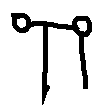} &
        \includegraphics[width=0.22\textwidth]{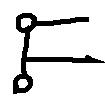} \\
        \small Image A & \small Image B
    \end{tabular}
    
    \vspace{0.6em}
    
    \textbf{Negative Example} \\
    \begin{tabular}{cc}
        \includegraphics[width=0.22\textwidth]{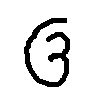} &
        \includegraphics[width=0.22\textwidth]{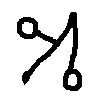} \\
        \small Image C & \small Image D
    \end{tabular}
    
    \caption{\textbf{In-context learning (ICL) prompting setup.} 
    The system prompt includes two labeled examples: a positive pair (top) where one image is a rotated version of the other, and a negative pair (bottom) where the images are not related by rotation.}
    
    \label{fig:icl_prompt}
\end{figure}

\clearpage
\newpage

\subsection{Rotational Grid Setup}\label{sec:supp-rot-grid}

\begin{figure}[h]
    \centering
    
    \fbox{
    \begin{minipage}{0.9\linewidth}
    \small
    \textbf{System Prompt:} \\
    You are a vision reasoning model. You will be shown ``Rotational Grids'' which display a character in four orientations simultaneously. The layout is: Top-Left: Original Image ($0^\circ$), Top-Right: Rotated $270^\circ$, Bottom-Left: Rotated $90^\circ$, Bottom-Right: Rotated $180^\circ$. Use the labels in the grid to understand the transformation.
    \end{minipage}
    }
    
    \vspace{0.8em}
    
    \textbf{Reference 1: Rotational Grid for Character A} \\
    \includegraphics[width=0.45\textwidth]{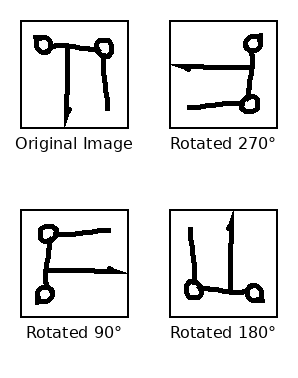}
    
    \smallskip
    {\small This grid defines the rotation patterns.}
    
    \vspace{0.8em}
    
    \textbf{Reference 2: Rotational Grid for Character B} \\
    \includegraphics[width=0.45\textwidth]{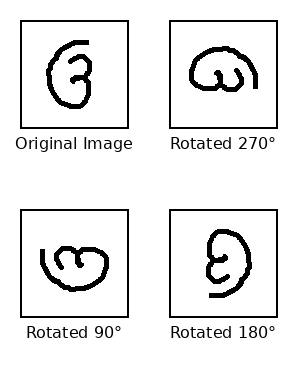}
    
    \smallskip
    {\small This grid defines the rotation patterns.}
    
    \caption{\textbf{Rotational grid prompting setup.} 
    The model is first given a structured system prompt describing the layout of rotational grids. 
    Two reference grids (Character A and Character B) are then provided to illustrate rotations across different characters.}
    
    \label{fig:rotation_grid_prompt}
\end{figure}

\clearpage
\newpage

\section{Additional Results}
\label{sec:supp-full}

\subsection{Rotation recognition performance across character datasets (additional)}
\label{sec:supp-rot-char}

\begin{table}[!h]
\centering
\tiny

\begin{tabular}{lcccccccccccc}
\toprule
& \multicolumn{3}{>{\columncolor{cyan!10}}c}{\textbf{\qwenSeven}} 
& \multicolumn{3}{>{\columncolor{cyan!10}}c}{\textbf{\qwenThirtyTwo}}
& \multicolumn{3}{>{\columncolor{cyan!10}}c}{\textbf{\qwenSeventyTwo}}
& \multicolumn{3}{>{\columncolor{blue!10}}c}{\textbf{\qwenThreeFour}} \\
\cmidrule(lr){2-4} \cmidrule(lr){5-7} \cmidrule(lr){8-10} \cmidrule(lr){11-13}

\textbf{Dataset} 
& \textbf{Acc.} & \textbf{TNR} & \textbf{TPR}
& \textbf{Acc.} & \textbf{TNR} & \textbf{TPR}
& \textbf{Acc.} & \textbf{TNR} & \textbf{TPR}
& \textbf{Acc.} & \textbf{TNR} & \textbf{TPR} \\
\midrule

Times New Roman
& \textcolor{teal}{50.96} & 100.00 & 01.92
& \textcolor{red}{54.13} & 100.00 & 08.27
& \textcolor{red}{52.40} & 100.00 & 04.81
& \textcolor{teal}{50.29} & 100.00 & 00.58 \\

Handwritten English
& 50.96 & 98.08 & 03.85
& \textcolor{teal}{62.50} & 98.08 & 26.92
& 57.21 & 100.00 & 14.42
& \textcolor{red}{50.00} & 100.00 & 00.00 \\

Omniglot
& \textcolor{red}{50.70} & 99.45 & 01.95
& 54.20 & 94.73 & 13.68
& \textcolor{teal}{57.48} & 96.51 & 18.46
& 50.24 & 99.98 & 00.51 \\

\midrule

& \multicolumn{3}{>{\columncolor{blue!10}}c}{\textbf{\qwenThreeEight}}
& \multicolumn{3}{>{\columncolor{blue!10}}c}{\textbf{\qwenThreeThirty}}
& \multicolumn{3}{>{\columncolor{gray!10}}c}{\textbf{\gpt}}
& \multicolumn{3}{>{\columncolor{gray!10}}c}{\textbf{\gemini}} \\
\cmidrule(lr){2-4} \cmidrule(lr){5-7} \cmidrule(lr){8-10} \cmidrule(lr){11-13}

\textbf{Dataset} 
& \textbf{Acc.} & \textbf{TNR} & \textbf{TPR}
& \textbf{Acc.} & \textbf{TNR} & \textbf{TPR}
& \textbf{Acc.} & \textbf{TNR} & \textbf{TPR}
& \textbf{Acc.} & \textbf{TNR} & \textbf{TPR} \\
\midrule

Times New Roman
& 50.19 & 100.00 & 00.38
& \textcolor{teal}{67.12} & 100.00 & 34.23
& 72.31 & 95.00 & 49.62
& \textcolor{teal}{89.23} & 100.00 & 78.46 \\

Handwritten English
& \textcolor{red}{50.00} & 100.00 & 00.00
& \textcolor{red}{56.44} & 100.00 & 12.88
& \textcolor{red}{67.02} & 96.35 & 37.69
& \textcolor{red}{67.40} & 99.42 & 35.38 \\

Omniglot
& \textcolor{teal}{50.99} & 99.96 & 02.03
& 56.55 & 99.76 & 13.34
& \textcolor{teal}{75.08} & 80.95 & 69.21
& 76.49 & 98.69 & 54.29 \\

\bottomrule
\end{tabular}

\caption{\textbf{Rotation recognition performance across character datasets.} 
Best and worst accuracy per model across datasets are highlighted in \textcolor{teal}{teal} and \textcolor{red}{red}, respectively. Performance aggregated over rotation angles $10^\circ$–$90^\circ$. 
While TNR remains near-perfect across all models, TPR is consistently low, 
indicating a failure to recognize rotated variants. Closed-source models perform 
better than open-source models, but the failure persists across all models. We additionally include results for {\qwenSeventyTwo} and {\qwenThreeFour}.}
\label{tab:rotation_results_app}
\end{table}










\subsection{Linear Probing of Rotation Information}
\label{sec:linear_probe}

We train a linear probe to further show that rotation information is linearly decodable from vision encoder features, yet downstream VLMs fail to utilize it.

We classify rotation angle into $9$ classes (10°, 20°, 30°, 40°, 50°, 60°, 70°, 80°, 90°) using a linear probe. For each image pair (I, $t_{rotation(I)}$) where $t_{rotation}$ denotes rotation by a fixed angle, we extract pooled encoder features from both the original image I and the rotated image $t_{rotation(I)}$ (Sec.~\ref{sec:trans_inv}). We concatenate these feature vectors to create a combined representation that capture both the original and rotated image information. This concatenated representation is passed to a linear classifier to predict the rotation angle.

We study this on all samples from the Omniglot dataset (Sec.~\ref{sec:datasets}) using a $60$/$20$/$20$ train/validation/test split stratified by script classes. Each vision encoder is independently probed with identical hyperparameters: learning rate = $1e-3$, batch size = $128$, AdamW optimizer, and trained for $200$ epochs. We report performance on the unseen test scripts.

\begin{table}[h]
    \centering
    \small
    \begin{tabular}{lccccccccc|c}
        \toprule
        Encoder & 10° & 20° & 30° & 40° & 50° & 60° & 70° & 80° & 90° & Overall \\
        \midrule
        SigLIP-SO400M-384 & 86.5 & 81.8 & 71.5 & 57.3 & 64.0 & 74.6 & 84.4 & 84.9 & 100.0 & 78.3 \\
        Qwen2.5-VL-7B & 94.3 & 86.1 & 78.4 & 68.3 & 64.1 & 77.2 & 87.7 & 93.5 & 99.8 & 83.3 \\
        DINOv2 ViT-L/14 & 98.0 & 96.3 & 94.7 & 83.2 & 82.4 & 92.6 & 95.6 & 97.3 & 99.7 & 93.3 \\
        CLIP ViT-L/14-336 & 91.3 & 86.2 & 79.0 & 69.8 & 68.1 & 79.8 & 85.4 & 89.4 & 99.1 & 83.1 \\
        \bottomrule
    \end{tabular}
    \caption{\textbf{Angle classification accuracy across vision encoders on Omniglot characters.} All encoders substantially exceed chance performance ($11.1$\%), suggesting that rotation information is linearly recoverable from vision encoder features.}
    \label{tab:rotation_probe}
\end{table}

All four vision encoders achieve accuracy well over random chance ($11.1$\%): SigLIP-SO400M ($78.3$\%), Qwen2.5-VL-7B ($83.3$\%), DINOv2 ($93.3$\%), CLIP ($83.1$\%) (Table~\ref{tab:rotation_probe}). This indicates that rotation angle appears to be linearly decodable from the vision encoder features, consistent with our analysis in Figure.~\ref{fig:encoder_graphs_omniglot}.

Yet as shown in Sec.~\ref{sec:rot}, downstream VLMs using these same encoders collapse: Qwen2.5-VL-7B achieves only $50.72$\% accuracy on Omniglot rotation, and Idefics2 (built on SigLIP-SO400M-384) achieves only $53.55$\% on Omniglot scripts (Appendix D).

\clearpage

\subsection{PACS rotation results for $10^\circ$ - $90^\circ$}
\label{sec:supp-rot-pacs-0-90}

\begin{table}[!h]
\resizebox{\textwidth}{!}{%
\begin{tabular}{lcccccccccccccccccc}
\toprule
& \multicolumn{3}{>{\columncolor{cyan!10}}c}{\textbf{\qwenSeven}} 
& \multicolumn{3}{>{\columncolor{cyan!10}}c}{\textbf{\qwenThirtyTwo}}
& \multicolumn{3}{>{\columncolor{blue!10}}c}{\textbf{\qwenThreeEight}}
& \multicolumn{3}{>{\columncolor{blue!10}}c}{\textbf{\qwenThreeThirty}}
& \multicolumn{3}{>{\columncolor{gray!10}}c}{\textbf{\gpt}}
& \multicolumn{3}{>{\columncolor{gray!10}}c}{\textbf{\gemini}} \\
\cmidrule(lr){2-4}
\cmidrule(lr){5-7}
\cmidrule(lr){8-10}
\cmidrule(lr){11-13}
\cmidrule(lr){14-16}
\cmidrule(lr){17-19}

\textbf{Domain} 
& \textbf{Acc.} & \textbf{TNR} & \textbf{TPR}
& \textbf{Acc.} & \textbf{TNR} & \textbf{TPR}
& \textbf{Acc.} & \textbf{TNR} & \textbf{TPR}
& \textbf{Acc.} & \textbf{TNR} & \textbf{TPR}
& \textbf{Acc.} & \textbf{TNR} & \textbf{TPR}
& \textbf{Acc.} & \textbf{TNR} & \textbf{TPR} \\
\midrule

Photo
& \textcolor{teal}{57.41} & 100.00 & 14.82
& \textcolor{teal}{83.73} & 100.00 & 67.45
& \textcolor{red}{52.09} & 100.00 & 4.18
& 82.14 & 100.00 & 64.27
& 98.91 & 99.95 & 97.86
& 94.70 & 100.00 & 89.41 \\

Art Painting
& 54.70 & 100.00 & 9.41
& 79.55 & 100.00 & 59.09
& \textcolor{teal}{52.91} & 100.00 & 5.82
& 77.23 & 100.00 & 54.45
& \textcolor{teal}{99.02} & 100.00 & 98.05
& \textcolor{teal}{95.80} & 100.00 & 91.59 \\

Cartoon
& 55.23 & 100.00 & 10.45
& 81.89 & 100.00 & 63.77
& 52.80 & 100.00 & 5.59
& \textcolor{teal}{82.50} & 100.00 & 65.00
& 97.95 & 100.00 & 95.91
& 93.23 & 100.00 & 86.45 \\

Sketch
& \textcolor{red}{53.36} & 100.00 & 6.73
& \textcolor{red}{73.07} & 100.00 & 46.14
& 52.66 & 100.00 & 5.32
& \textcolor{red}{76.00} & 100.00 & 52.00
& \textcolor{red}{95.61} & 99.64 & 91.59
& \textcolor{red}{90.30} & 99.86 & 80.73 \\

\noalign{\vskip 3pt}
\cdashline{1-19}
\noalign{\vskip 2pt}

Random guess 
& 50.00 & 50.00 & 50.00
& 50.00 & 50.00 & 50.00
& 50.00 & 50.00 & 50.00
& 50.00 & 50.00 & 50.00
& 50.00 & 50.00 & 50.00
& 50.00 & 50.00 & 50.00 \\

\bottomrule
\end{tabular}}

\vspace{-0.4em}
\caption{\textbf{Rotation recognition performance across PACS domains ($10^\circ$ – $90^\circ$).} Best and worst accuracy per model across domains are highlighted in \textcolor{teal}{teal} and \textcolor{red}{red}, respectively. Performance aggregated over rotations from $10^\circ$ to $90^\circ$ in increments of $10^\circ$. While TNR remains near-perfect across all models, TPR varies significantly across domains, with strong performance on photographs and substantial degradation on sketches, indicating limited geometric robustness.}
\vspace{-0.4em}
\label{tab:pacs_results_10_90_app}
\end{table}

\subsection{PACS identity experiment results}
\label{sec:supp-identity-pacs}

\begin{table}[!h]
\resizebox{\textwidth}{!}{%
\begin{tabular}{lcccccccccccccccccc}
\toprule
& \multicolumn{3}{>{\columncolor{cyan!10}}c}{\textbf{\qwenSeven}} 
& \multicolumn{3}{>{\columncolor{cyan!10}}c}{\textbf{\qwenThirtyTwo}}
& \multicolumn{3}{>{\columncolor{blue!10}}c}{\textbf{\qwenThreeEight}}
& \multicolumn{3}{>{\columncolor{blue!10}}c}{\textbf{\qwenThreeThirty}}
& \multicolumn{3}{>{\columncolor{gray!10}}c}{\textbf{\gpt}}
& \multicolumn{3}{>{\columncolor{gray!10}}c}{\textbf{\gemini}} \\
\cmidrule(lr){2-4}
\cmidrule(lr){5-7}
\cmidrule(lr){8-10}
\cmidrule(lr){11-13}
\cmidrule(lr){14-16}
\cmidrule(lr){17-19}

\textbf{Domain} 
& \textbf{Acc.} & \textbf{TNR} & \textbf{TPR}
& \textbf{Acc.} & \textbf{TNR} & \textbf{TPR}
& \textbf{Acc.} & \textbf{TNR} & \textbf{TPR}
& \textbf{Acc.} & \textbf{TNR} & \textbf{TPR}
& \textbf{Acc.} & \textbf{TNR} & \textbf{TPR}
& \textbf{Acc.} & \textbf{TNR} & \textbf{TPR} \\
\midrule

Photo
& 53.50 & 100.00 & 7.00
& \textcolor{teal}{100.00} & 100.00 & 100.00
& \textcolor{teal}{100.00} & 100.00 & 100.00
& \textcolor{teal}{100.00} & 100.00 & 100.00
& \textcolor{teal}{100.00} & 100.00 & 100.00
& \textcolor{teal}{100.00} & 100.00 & 100.00 \\

Art Painting
& \textcolor{red}{52.75} & 100.00 & 5.50
& \textcolor{teal}{100.00} & 100.00 & 100.00
& \textcolor{teal}{100.00} & 100.00 & 100.00
& \textcolor{teal}{100.00} & 100.00 & 100.00
& \textcolor{teal}{100.00} & 100.00 & 100.00
& \textcolor{teal}{100.00} & 100.00 & 100.00 \\

Cartoon
& 76.75 & 100.00 & 53.50
& \textcolor{teal}{100.00} & 100.00 & 100.00
& \textcolor{teal}{100.00} & 100.00 & 100.00
& \textcolor{teal}{100.00} & 100.00 & 100.00
& \textcolor{red}{99.75} & 100.00 & 99.50
& \textcolor{teal}{100.00} & 100.00 & 100.00 \\

Sketch
& \textcolor{teal}{92.25} & 100.00 & 84.50
& \textcolor{teal}{100.00} & 100.00 & 100.00
& \textcolor{teal}{100.00} & 100.00 & 100.00
& \textcolor{red}{99.75} & 100.00 & 99.50
& \textcolor{teal}{100.00} & 100.00 & 100.00
& \textcolor{teal}{100.00} & 100.00 & 100.00 \\

\noalign{\vskip 3pt}
\cdashline{1-19}
\noalign{\vskip 2pt}

Random guess 
& 50.00 & 50.00 & 50.00
& 50.00 & 50.00 & 50.00
& 50.00 & 50.00 & 50.00
& 50.00 & 50.00 & 50.00
& 50.00 & 50.00 & 50.00
& 50.00 & 50.00 & 50.00 \\

\bottomrule
\end{tabular}}

\vspace{-0.4em}
\caption{\textbf{Model performance for the identity transformation on PACS.} Best accuracy values per model across domains are highlighted in \textcolor{teal}{teal}, including ties, while worst values are shown in \textcolor{red}{red}. All models achieve near-perfect performance across domains, \textbf{except {\qwenSeven}}, which shows lower TPR on Photo and Art despite perfect TNR.}
\vspace{-0.4em}
\label{tab:pacs_identity_results_app}
\end{table}

\clearpage
\newpage
\subsection{Model performance with In-Context Learning}
\label{sec:supp-icl}

\begin{table}[!h]
\centering

\begin{subtable}[t]{\textwidth}
\begin{adjustbox}{width=\textwidth}
\begin{tabular}{lcccccccccccc}
\toprule
&
\multicolumn{2}{>{\columncolor{cyan!10}}c}{\textbf{\qwenSeven}}
& \multicolumn{2}{>{\columncolor{cyan!10}}c}{\textbf{\qwenThirtyTwo}}
& \multicolumn{2}{>{\columncolor{blue!10}}c}{\textbf{\qwenThreeEight}}
& \multicolumn{2}{>{\columncolor{blue!10}}c}{\textbf{\qwenThreeThirty}}
& \multicolumn{2}{>{\columncolor{gray!10}}c}{\textbf{\gpt}}
& \multicolumn{2}{>{\columncolor{gray!10}}c}{\textbf{\gemini}} \\
\cmidrule(lr){2-3}
\cmidrule(lr){4-5}
\cmidrule(lr){6-7}
\cmidrule(lr){8-9}
\cmidrule(lr){10-11}
\cmidrule(lr){12-13}
\textbf{ICL Setting}
& \textbf{TNR} & \textbf{\tprhead}
& \textbf{TNR} & \textbf{\tprhead}
& \textbf{TNR} & \textbf{\tprhead}
& \textbf{TNR} & \textbf{\tprhead}
& \textbf{TNR} & \textbf{\tprhead}
& \textbf{TNR} & \textbf{\tprhead} \\
\midrule
None
& 100.00 & 00.00\same
& 97.87 & 06.38\same
& 100.00 & 00.00\same
& 100.00 & 02.13\same
& 93.62 & 34.04\same
& 100.00 & 38.30\same \\
Few-shot
& 100.00 & 00.00\same
& 65.96 & \best{51.06}\up
& 100.00 & 34.04\up
& 78.72 & 72.34\up
& 91.49 & 85.11\up
& 97.87 & \textcolor{red}{34.04}\down \\
Rotational Grid
& 100.00 & 00.00\same
& 91.49 & 12.77\up
& 65.96 & \best{65.96}\up
& 23.40 & \best{87.23}\up
& 72.34 & \best{97.87}\up
& 93.62 & \best{48.94}\up \\
\bottomrule
\end{tabular}
\label{tab:table1_a}
\end{adjustbox}
\vspace{-0.05in}
\caption{Malayalam (top-tier script)}
\vspace{0.1in}
\end{subtable}

\begin{subtable}[t]{\textwidth}
\begin{adjustbox}{width=\textwidth}
\begin{tabular}{lcccccccccccc}
\toprule
&
\multicolumn{2}{>{\columncolor{cyan!10}}c}{\textbf{\qwenSeven}}
& \multicolumn{2}{>{\columncolor{cyan!10}}c}{\textbf{\qwenThirtyTwo}}
& \multicolumn{2}{>{\columncolor{blue!10}}c}{\textbf{\qwenThreeEight}}
& \multicolumn{2}{>{\columncolor{blue!10}}c}{\textbf{\qwenThreeThirty}}
& \multicolumn{2}{>{\columncolor{gray!10}}c}{\textbf{\gpt}}
& \multicolumn{2}{>{\columncolor{gray!10}}c}{\textbf{\gemini}} \\
\cmidrule(lr){2-3}
\cmidrule(lr){4-5}
\cmidrule(lr){6-7}
\cmidrule(lr){8-9}
\cmidrule(lr){10-11}
\cmidrule(lr){12-13}
\textbf{ICL Setting}
& \textbf{TNR} & \textbf{\tprhead}
& \textbf{TNR} & \textbf{\tprhead}
& \textbf{TNR} & \textbf{\tprhead}
& \textbf{TNR} & \textbf{\tprhead}
& \textbf{TNR} & \textbf{\tprhead}
& \textbf{TNR} & \textbf{\tprhead} \\
\midrule
None
& 100.00 & 00.00\same
& 100.00 & 13.95\same
& 100.00 & 00.00\same
& 100.00 & 00.00\same
& 88.37 & 37.21\same
& 97.67 & \best{51.16}\same \\
Few-shot
& 100.00 & 00.00\same
& 67.44 & \best{58.14}\up
& 97.67 & 09.30\up
& 76.74 & 67.44\up
& 74.42 & 86.05\up
& 83.72 & \textcolor{red}{34.88}\down \\
Rotational Grid
& 100.00 & 00.00\same
& 93.02 & \textcolor{red}{09.30}\down
& 81.40 & \best{60.47}\up
& 34.88 & \best{88.37}\up
& 53.49 & \best{90.70}\up
& 83.72 & \textcolor{red}{39.53}\down \\
\bottomrule
\end{tabular}
\label{tab:table1_b}
\end{adjustbox}
\vspace{-0.05in}
\caption{Grantha (top-tier script)}
\vspace{0.1in}
\end{subtable}

\begin{subtable}[t]{\textwidth}
\begin{adjustbox}{width=\textwidth}
\begin{tabular}{lcccccccccccc}
\toprule
&
\multicolumn{2}{>{\columncolor{cyan!10}}c}{\textbf{\qwenSeven}}
& \multicolumn{2}{>{\columncolor{cyan!10}}c}{\textbf{\qwenThirtyTwo}}
& \multicolumn{2}{>{\columncolor{blue!10}}c}{\textbf{\qwenThreeEight}}
& \multicolumn{2}{>{\columncolor{blue!10}}c}{\textbf{\qwenThreeThirty}}
& \multicolumn{2}{>{\columncolor{gray!10}}c}{\textbf{\gpt}}
& \multicolumn{2}{>{\columncolor{gray!10}}c}{\textbf{\gemini}} \\
\cmidrule(lr){2-3}
\cmidrule(lr){4-5}
\cmidrule(lr){6-7}
\cmidrule(lr){8-9}
\cmidrule(lr){10-11}
\cmidrule(lr){12-13}
\textbf{ICL Setting}
& \textbf{TNR} & \textbf{\tprhead}
& \textbf{TNR} & \textbf{\tprhead}
& \textbf{TNR} & \textbf{\tprhead}
& \textbf{TNR} & \textbf{\tprhead}
& \textbf{TNR} & \textbf{\tprhead}
& \textbf{TNR} & \textbf{\tprhead} \\
\midrule
None
& 100.00 & 00.00\same
& 88.00 & 16.00\same
& 100.00 & 00.00\same
& 100.00 & 00.00\same
& 100.00 & 20.00\same
& 96.00 & 32.00\same \\
Few-shot
& 100.00 & 00.00\same
& 68.00 & \best{76.00}\up
& 100.00 & 16.00\up
& 68.00 & 72.00\up
& 80.00 & 68.00\up
& 92.00 & 32.00\same \\
Rotational Grid
& 100.00 & 00.00\same
& 88.00 & 24.00\up
& 72.00 & \best{88.00}\up
& 40.00 & \best{76.00}\up
& 60.00 & \best{88.00}\up
& 88.00 & \best{56.00}\up \\
\bottomrule
\end{tabular}
\label{tab:table1_c}
\end{adjustbox}
\vspace{-0.05in}
\caption{Tengwar (medium-tier script)}
\vspace{0.1in}
\end{subtable}

\begin{subtable}[t]{\textwidth}
\begin{adjustbox}{width=\textwidth}
\begin{tabular}{lcccccccccccc}
\toprule
&
\multicolumn{2}{>{\columncolor{cyan!10}}c}{\textbf{\qwenSeven}}
& \multicolumn{2}{>{\columncolor{cyan!10}}c}{\textbf{\qwenThirtyTwo}}
& \multicolumn{2}{>{\columncolor{blue!10}}c}{\textbf{\qwenThreeEight}}
& \multicolumn{2}{>{\columncolor{blue!10}}c}{\textbf{\qwenThreeThirty}}
& \multicolumn{2}{>{\columncolor{gray!10}}c}{\textbf{\gpt}}
& \multicolumn{2}{>{\columncolor{gray!10}}c}{\textbf{\gemini}} \\
\cmidrule(lr){2-3}
\cmidrule(lr){4-5}
\cmidrule(lr){6-7}
\cmidrule(lr){8-9}
\cmidrule(lr){10-11}
\cmidrule(lr){12-13}
\textbf{ICL Setting}
& \textbf{TNR} & \textbf{\tprhead}
& \textbf{TNR} & \textbf{\tprhead}
& \textbf{TNR} & \textbf{\tprhead}
& \textbf{TNR} & \textbf{\tprhead}
& \textbf{TNR} & \textbf{\tprhead}
& \textbf{TNR} & \textbf{\tprhead} \\
\midrule
None
& 95.83 & 00.00\same
& 91.67 & 16.67\same
& 100.00 & 00.00\same
& 100.00 & 00.00\same
& 87.50 & 29.17\same
& 100.00 & 20.83\same \\
Few-shot
& 100.00 & 00.00\same
& 58.33 & \best{45.83}\up
& 100.00 & 12.50\up
& 50.00 & 95.83\up
& 83.33 & 66.67\up
& 91.67 & 25.00\up \\
Rotational Grid
& 100.00 & 00.00\same
& 100.00 & 20.83\up
& 33.33 & \best{83.33}\up
& 04.17 & \best{100.00}\up
& 54.17 & \best{87.50}\up
& 83.33 & \best{58.33}\up \\
\bottomrule
\end{tabular}
\label{tab:table1_d}
\end{adjustbox}
\vspace{-0.05in}
\caption{Balinese (medium-tier script)}
\vspace{0.1in}
\end{subtable}

\begin{subtable}[t]{\textwidth}
\begin{adjustbox}{width=\textwidth}
\begin{tabular}{lcccccccccccc}
\toprule
&
\multicolumn{2}{>{\columncolor{cyan!10}}c}{\textbf{\qwenSeven}}
& \multicolumn{2}{>{\columncolor{cyan!10}}c}{\textbf{\qwenThirtyTwo}}
& \multicolumn{2}{>{\columncolor{blue!10}}c}{\textbf{\qwenThreeEight}}
& \multicolumn{2}{>{\columncolor{blue!10}}c}{\textbf{\qwenThreeThirty}}
& \multicolumn{2}{>{\columncolor{gray!10}}c}{\textbf{\gpt}}
& \multicolumn{2}{>{\columncolor{gray!10}}c}{\textbf{\gemini}} \\
\cmidrule(lr){2-3}
\cmidrule(lr){4-5}
\cmidrule(lr){6-7}
\cmidrule(lr){8-9}
\cmidrule(lr){10-11}
\cmidrule(lr){12-13}
\textbf{ICL Setting}
& \textbf{TNR} & \textbf{\tprhead}
& \textbf{TNR} & \textbf{\tprhead}
& \textbf{TNR} & \textbf{\tprhead}
& \textbf{TNR} & \textbf{\tprhead}
& \textbf{TNR} & \textbf{\tprhead}
& \textbf{TNR} & \textbf{\tprhead} \\
\midrule
None
& 100.00 & 00.00\same
& 65.38 & 15.38\same
& 100.00 & 00.00\same
& 100.00 & 03.85\same
& 100.00 & 07.69\same
& 100.00 & \best{50.00}\same \\
Few-shot
& 100.00 & 00.00\same
& 88.46 & \best{19.23}\up
& 100.00 & \best{11.54}\up
& 92.31 & \best{50.00}\up
& 92.31 & \best{73.08}\up
& 96.15 & \textcolor{red}{26.92}\down \\
Rotational Grid
& 100.00 & 00.00\same
& 100.00 & \textcolor{red}{00.00}\down
& 100.00 & \best{11.54}\up
& 73.08 & \best{50.00}\up
& 69.23 & 65.38\up
& 100.00 & \textcolor{red}{46.15}\down \\
\bottomrule
\end{tabular}
\label{tab:table1_e}
\end{adjustbox}
\vspace{-0.05in}
\caption{Braille (low-tier script)}
\vspace{0.1in}
\end{subtable}

\begin{subtable}[t]{\textwidth}
\begin{adjustbox}{width=\textwidth}
\begin{tabular}{lcccccccccccc}
\toprule
&
\multicolumn{2}{>{\columncolor{cyan!10}}c}{\textbf{\qwenSeven}}
& \multicolumn{2}{>{\columncolor{cyan!10}}c}{\textbf{\qwenThirtyTwo}}
& \multicolumn{2}{>{\columncolor{blue!10}}c}{\textbf{\qwenThreeEight}}
& \multicolumn{2}{>{\columncolor{blue!10}}c}{\textbf{\qwenThreeThirty}}
& \multicolumn{2}{>{\columncolor{gray!10}}c}{\textbf{\gpt}}
& \multicolumn{2}{>{\columncolor{gray!10}}c}{\textbf{\gemini}} \\
\cmidrule(lr){2-3}
\cmidrule(lr){4-5}
\cmidrule(lr){6-7}
\cmidrule(lr){8-9}
\cmidrule(lr){10-11}
\cmidrule(lr){12-13}
\textbf{ICL Setting}
& \textbf{TNR} & \textbf{\tprhead}
& \textbf{TNR} & \textbf{\tprhead}
& \textbf{TNR} & \textbf{\tprhead}
& \textbf{TNR} & \textbf{\tprhead}
& \textbf{TNR} & \textbf{\tprhead}
& \textbf{TNR} & \textbf{\tprhead} \\
\midrule
None
& 100.00 & 00.00\same
& 100.00 & 31.03\same
& 100.00 & 03.45\same
& 100.00 & 00.00\same
& 96.55 & 27.59\same
& 100.00 & 51.72\same \\
Few-shot
& 100.00 & \best{03.45}\up
& 72.41 & \best{89.66}\up
& 100.00 & 41.38\up
& 100.00 & 62.07\up
& 100.00 & 86.21\up
& 100.00 & \textcolor{red}{27.59}\down \\
Rotational Grid
& 100.00 & 00.00\same
& 100.00 & 31.03\same
& 86.21 & \best{86.21}\up
& 75.86 & \best{75.86}\up
& 86.21 & \best{96.55}\up
& 93.10 & \best{72.41}\up \\
\bottomrule
\end{tabular}
\label{tab:table1_f}
\end{adjustbox}
\vspace{-0.05in}
\caption{Anglo-Saxon Futhorc (low-tier script)}
\vspace{0.1in}
\end{subtable}

\caption{\textbf{Model performance with In-Context Learning}}
\label{tab:icl_final_app}
\end{table}

\clearpage

\subsection{Investigating Response Bias and Prompt Sensitivity}
\label{sec:prompt_analysis}

To determine whether performance degradation on the rotation task stems from a strong ``No'' bias or a genuine lack of geometric reasoning, we evaluated models using three different prompt variants:

\begin{itemize}
    \item \textbf{Original:} ``If I rotate the first image, can I get the second image? Answer in curly brackets, e.g. \{Yes\} or \{No\}.''
    \item \textbf{Balanced:} ``Are these two images rotated versions of each other?'' We denote this as balanced as it removes the directional asymmetry present in the Original prompt.
    \item \textbf{Forced-choice:} We frame the task as a multiple-choice question, using the prompt: ``Examine these two images carefully. Which statement best describes these two images? (A) Both images show the same character, but one is a rotated version of the other. (B) Both images show different characters.''
\end{itemize}

\begin{table}[h]
    \centering
    \small
    \begin{tabular}{llccc}
        \toprule
        Dataset & Prompt Variant & Accuracy & TPR & TNR \\
        \midrule
        Times New Roman & Original & 52.56 & 5.98 & 99.15 \\
        & Balanced & 51.71 & 3.42 & 100.0 \\
        & Forced Choice & 90.49 & 80.98 & 100.0 \\
        Handwritten English & Original & 61.22 & 32.05 & 90.38 \\
        & Balanced & 61.97 & 23.93 & 100.0 \\
        & Forced Choice & 76.60 & 99.36 & 53.85 \\
        Omniglot & Original & 52.00 & 13.47 & 94.93 \\
        & Balanced & 53.10 & 14.23 & 96.41 \\
        & Forced Choice & 65.30 & 88.80 & 39.11 \\
        \bottomrule
    \end{tabular}
    \caption{\textbf{Performance metrics across datasets and prompt variants for Qwen2.5-VL-32B.} Forced-choice MCQ dramatically improves TPR (+ 65-75 \%) on all datasets, demonstrating strong response bias. However, TNR collapse on Handwritten English (53.85\%) and Omniglot (39.11\%) indicates the model lacks stable geometric reasoning -- it trades false negatives for false positives rather than achieving robust rotation understanding.}
    \label{tab:qwen_prompt}
\end{table}

\begin{table}[h]
    \centering
    \small
    \begin{tabular}{llccc}
        \toprule
        Dataset & Prompt Variant & Accuracy & TPR & TNR \\
        \midrule
        Times New Roman & Original & 89.32 & 78.6 & 100.0 \\
        & Balanced & 88.1 & 76.3 & 100.0 \\
        & Forced Choice & 87.9 & 75.9 & 100.0 \\
        Handwritten English & Original & 68.3 & 37.0 & 99.6 \\
        & Balanced & 69.6 & 39.5 & 99.6 \\
        & Forced Choice & 67.2 & 34.4 & 100.0 \\
        Omniglot & Original & 76.9 & 55.1 & 98.7 \\
        & Balanced & 75.5 & 54.5 & 98.9 \\
        & Forced Choice & 73.3 & 50.5 & 98.7 \\
        \bottomrule
    \end{tabular}
    \caption{\textbf{Performance metrics across datasets and prompt variants for Gemini-2.5-Pro.} Different prompt formats shows consistent performance across all datasets, demonstrating that the drop in performance in Handwritten English and Omniglot is not due to prompt sensitivity.}
    \label{tab:gemini_prompt}
\end{table}

For Qwen2.5-VL-32B~\citep{qwen2}, the balanced prompt shows no improvement ($1$\%), and while forced-choice improves TPR (+$65$--$75$\%), it simultaneously collapses TNR from $94.93$\% to $39.11$\% on Omniglot -- the model trades false negatives for false positives rather than achieving robust geometric understanding. In contrast, Gemini-2.5-Pro~\citep{gemini} shows no response bias, accuracy remains stable across all three variants ($2$--$4$\%), demonstrating that its failure is a fundamental geometric limitation, not a prompting artifact.

\clearpage

\subsection{Human Baseline Study}
\label{sec:human_study}

We conducted a human baseline study on a stratified subset of our benchmark. Nine participants were given $60$ image pairs across $5$ scripts: Times New Roman, Handwritten English, Armenian, Greek, and Malayalam. Each trial mimics the rotation experiment in Sec.~\ref{sec:rot}: participants are shown either an image and its rotated counterpart (positive samples, with rotation angles uniformly sampled from $\{10^\circ, 20^\circ, \dots, 90^\circ\}$) or two different characters (negative samples), and asked the same prompt: ``If I rotate the first image, can I get the second image? Answer 'Yes' or 'No'.''

\begin{table}[h]
    \centering
    \small
    \begin{tabular}{lcc}
        \toprule
        Category & Accuracy & Avg Time \\
        \midrule
        Armenian & 100\% & 2.78s $\pm$ 1.13s \\
        Greek & 100\% & 2.70s $\pm$ 1.12s \\
        Hand-English & 100\% & 2.46s $\pm$ 1.08s \\
        Malayalam & 100\% & 2.91s $\pm$ 1.29s \\
        Times New Roman & 100\% & 2.28s $\pm$ 1.05s \\
        \bottomrule
    \end{tabular}
    \caption{\textbf{Human baseline performance on the rotation task by script category.} Humans maintain perfect geometric reasoning regardless of script familiarity, highlighting a fundamental vulnerability in current VLMs.}
    \label{tab:human_baseline}
\end{table}

From this human study, we conclude that:

\begin{itemize}
    \item Humans achieved perfect accuracy ($100$\%) across all scripts (Table \ref{tab:human_baseline}), establishing a clear performance ceiling and confirming the rotation task is trivially easy for human vision.
    \item Response times were only marginally longer for unfamiliar Omniglot scripts (Malayalam: 2.91s, Armenian: 2.78s) than for familiar Latin text (Times New Roman: 2.28s). This minor slowdown does indicate perceptual familiarity with known scripts but does not indicate task difficulty.
\end{itemize}

Thus, we conclude that any model failures reported on this task reflect a genuine vision limitation in VLMs, not task ambiguity.

\subsection{Statistical Robustness and Confidence Intervals}
\label{sec:stats_robustness}

To ensure the statistical significance of our findings, we report 95\% bootstrap confidence intervals (10,000 resamples) for all models across the rotation and scale tasks in Tables~\ref{tab:rotation_ci} and~\ref{tab:scale_ci}.

\begin{table}[h]
    \centering
    \scriptsize
    \resizebox{\textwidth}{!}{
    \begin{tabular}{lcccccc}
        \toprule
        Dataset & Qwen2.5-VL-7B & Qwen2.5-VL-32B & Qwen3-VL-8B & Qwen3-VL-30B & GPT-5.2 & Gemini-2.5-Pro \\
        \midrule
        Times New Roman & 51.07 [47.75, 54.17] & 52.67 [49.36, 55.67] & 50.11 [46.79, 53.21] & 65.81 [62.50, 68.80] & 74.25 [71.37, 76.92] & 89.32 [87.18, 91.24] \\
        Handwritten English & 50.85 [47.86, 54.17] & 62.50 [59.40, 65.60] & 50.00 [47.01, 53.42] & 55.98 [52.99, 59.19] & 67.84 [64.53, 70.94] & 68.27 [65.49, 71.26] \\
        Omniglot & 50.72 [50.14, 51.25] & 54.17 [53.61, 54.74] & 51.01 [50.48, 51.56] & 56.64 [56.10, 57.20] & 75.55 [75.04, 76.02] & 76.90 [76.40, 77.36] \\
        \bottomrule
    \end{tabular}
    }
    \caption{\textbf{Rotation task: Accuracy with 95\% bootstrap confidence intervals (10,000 resamples).} Narrow intervals confirm that the reported results are statistically robust.}
    \label{tab:rotation_ci}
\end{table}

\begin{table}[h]
    \centering
    \scriptsize
    \resizebox{\textwidth}{!}{
    \begin{tabular}{lcccccc}
        \toprule
        Dataset & Qwen2.5-VL-7B & Qwen2.5-VL-32B & Qwen3-VL-8B & Qwen3-VL-30B & GPT-5.2 & Gemini-2.5-Pro \\
        \midrule
        Times New Roman & 98.80 [97.60, 99.76] & 99.76 [99.28, 100.00] & 100.00 [100.00, 100.00] & 98.79 [97.60, 99.76] & 98.79 [97.60, 99.76] & 99.51 [98.56, 100.00] \\
        Handwritten English & 96.77 [95.16, 98.19] & 95.36 [93.54, 97.18] & 97.59 [95.19, 99.52] & 97.59 [96.15, 98.80] & 98.07 [96.63, 99.28] & 96.63 [94.23, 99.04] \\
        Omniglot & 77.40 [76.68, 78.16] & 74.21 [73.48, 74.92] & 76.05 [75.34, 76.79] & 77.04 [76.32, 77.80] & 79.72 [79.00, 80.42] & 82.56 [81.91, 83.21] \\
        \bottomrule
    \end{tabular}
    }
    \caption{\textbf{Scale task: Accuracy with 95\% bootstrap confidence intervals (10,000 resamples).} Narrow intervals confirm that the reported results are statistically robust.}
    \label{tab:scale_ci}
\end{table}

We report 95\% bootstrap confidence intervals (10,000 resamples) for all models across datasets in Tables~\ref{tab:rotation_ci} and~\ref{tab:scale_ci}. The confidence intervals are narrow across all models, datasets, and tasks, and the performance gaps between familiar and unfamiliar scripts are substantially larger than the interval widths (e.g., Gemini-2.5-Pro for the rotation task: 89.32\% [87.18, 91.24] on Times New Roman vs. 76.90\% [76.40, 77.36] on Omniglot). This confirms that the reported differences are statistically meaningful.

\subsection{Evaluating Additional Open-Source Models}
\label{sec:additional_models}

We evaluate two additional open-source model families: Molmo2-8B~\citep{clark2026molmo2} and InternVL3.5-8B~\citep{wang2025internvl3}. Both models show that geometric reasoning failures persist across architectures. On the scale-invariance task (Table~\ref{tab:scale_additional}), both follow the semantic-richness degradation pattern observed in Qwen variants, with TPR dropping from Times New Roman to Omniglot scripts (Molmo2-8B: 50.96\% $\to$ 12.52\%; InternVL3.5-8B: 96.15\% $\to$ 29.22\%). On rotation (Table~\ref{tab:rotation_additional}), each model exhibits a distinct but equally revealing failure mode: Molmo2-8B shows strong ``No'' bias with high TNR ($\sim$97\%) but near-zero TPR on Times New Roman (5.13\%), while InternVL3.5-8B shows the opposite -- high TPR (88--100\%) but TNR collapses across all datasets (Times New Roman: 16.03\%, Handwritten English: 31.41\%, Omniglot: 41.17\%). Neither model demonstrates robust geometric reasoning capabilities. These results across three distinct model families (Qwen~\citep{qwen2}, Molmo~\citep{clark2026molmo2}, InternVL~\citep{wang2025internvl3}) demonstrate that the vulnerability is not architecture-specific but generalizes across current VLMs.

\begin{table}[h]
    \centering
    \scriptsize
    \resizebox{\textwidth}{!}{
    \begin{tabular}{lcccccc}
        \toprule
        Dataset & Molmo2-8B Acc. & Molmo2-8B TNR & Molmo2-8B TPR & InternVL3.5-8B Acc. & InternVL3.5-8B TNR & InternVL3.5-8B TPR \\
        \midrule
        Times New Roman & 75.48 & 100.00 & 50.96 & 98.07 & 100.00 & 96.15 \\
        Handwritten English & 73.79 & 100.00 & 47.59 & 82.69 & 100.00 & 65.38 \\
        Omniglot & 56.07 & 99.61 & 12.52 & 64.28 & 99.35 & 29.22 \\
        Random guess & 50.00 & 50.00 & 50.00 & 50.00 & 50.00 & 50.00 \\
        \bottomrule
    \end{tabular}
    }
    \caption{\textbf{Model performance on the scale-invariance task aggregated across all scales for open-sourced models Molmo2-8B and InternVL3.5-8B.}}
    \label{tab:scale_additional}
\end{table}

\begin{table}[h]
    \centering
    \scriptsize
    \resizebox{\textwidth}{!}{
    \begin{tabular}{lcccccc}
        \toprule
        Dataset & Molmo2-8B Acc. & Molmo2-8B TNR & Molmo2-8B TPR & InternVL3.5-8B Acc. & InternVL3.5-8B TNR & InternVL3.5-8B TPR \\
        \midrule
        Times New Roman & 51.60 & 98.08 & 5.13 & 58.01 & 16.03 & 100.00 \\
        Handwritten English & 70.41 & 96.15 & 44.66 & 62.82 & 31.41 & 94.23 \\
        Omniglot & 69.09 & 96.92 & 41.27 & 64.83 & 41.17 & 88.48 \\
        Random guess & 50.00 & 50.00 & 50.00 & 50.00 & 50.00 & 50.00 \\
        \bottomrule
    \end{tabular}
    }
    \caption{\textbf{Model performance on the rotation task aggregated across all angles for open-sourced models Molmo2-8B and InternVL3.5-8B.}}
    \label{tab:rotation_additional}
\end{table}

\subsection{The Role of Data Domain and Semantic Grounding}
\label{sec:data_domain_grounding}

To investigate whether performance degradation is a consequence of distribution shift from the training domain, we explore if domain familiarity correlates with geometric reasoning ability. Specifically, we test whether fine-tuning a model on a particular data domain using generic, non-geometric semantic tasks enables it to better access the latent geometric structure already present in its vision encoder.

We LoRA fine-tune Qwen2.5-VL-7B on 3,729 PACS sketch images (completely separate from the 200 test sketch images described in Sec.~\ref{sec:datasets}). Crucially, we construct Visual Question Answering (VQA) pairs using purely semantic descriptions (e.g., Prompt: ``Describe the image.'' $\to$ Response: ``The image shows a sketch of a dog.''). No geometric supervision, rotation examples, or scale information were provided during this phase. Throughout this process, the vision encoder remained frozen; only the language decoder was updated via LoRA for 3 epochs. We evaluate both the baseline and the fine-tuned models on the rotation and scale tasks using the held-out 200 PACS sketch images.

\begin{table}[h]
    \centering
    \small
    \begin{tabular}{llccc}
        \toprule
        Model & Task & Acc. & TPR & TNR \\
        \midrule
        Baseline & Rotation & 56.36 & 12.72 & 100.00 \\
        Fine-tuned & Rotation & 90.75 & 81.50 & 100.00 \\
        \midrule
        Baseline & Scale & 92.94 & 86.62 & 99.25 \\
        Fine-tuned & Scale & 96.56 & 93.75 & 99.38 \\
        \bottomrule
    \end{tabular}
    \caption{\textbf{LoRA fine-tuning Qwen2.5-VL-7B on semantic VQA pairs (no geometric supervision) of sketch images from PACS improves rotation recognition.} Rotation TPR improves by +68.78\% while TNR is preserved.}
    \label{tab:lora_pacs_results}
\end{table}

As shown in Table~\ref{tab:lora_pacs_results}, although the fine-tuning data contained zero rotation or scale examples, the rotation TPR improved dramatically from 12.72\% to 81.50\% (+68.78\%). This finding directly validates our core hypothesis: while the frozen vision encoder faithfully preserves geometric features (Sec.~\ref{sec:enc_rot}), the language decoder struggles to access it without semantic grounding. Fine-tuning the language decoder alone with semantic labels, without modifying the visual encoder or providing any geometric supervision - substantially improves geometric reasoning on sketch images.

This finding is corroborated by concurrent work~\cite{shahgir2026vlms}, where arbitrary semantic labels are assigned to novel shapes (e.g., naming a random squiggle ``John'') and used to fine-tune VLMs, improving performance on visual correspondence tasks for these visual inputs.
\end{document}